%% file: main.tex
\documentclass[letterpaper, 10 pt, conference]{IEEEtran}

\IEEEoverridecommandlockouts                              

\usepackage{hyperref}
\usepackage[utf8]{inputenc}
\usepackage[T1]{fontenc}
\usepackage[dvipsnames]{xcolor}
\usepackage{dirtree}
\usepackage{makecell}
\usepackage{blindtext}
\usepackage{cite}
\usepackage{subfigure}

\usepackage{graphicx}

\usepackage{bm}
\usepackage{mathtools}
\usepackage{amsmath}

\usepackage{multicol}
\usepackage{multirow}
\usepackage{makecell}
\usepackage{array}
\usepackage{arydshln}

\usepackage{kotex}
\usepackage{cleveref}
\usepackage{amssymb}
\usepackage{pifont}
\usepackage[switch,columnwise]{lineno}

\definecolor{color1}{rgb}{0.0, 0.0, 0.0}
\definecolor{color2}{rgb}{0.0, 0.0, 0.0}
\definecolor{color3}{rgb}{0.0, 0.0, 0.0}

\def\Tau{{\rm T}}

\title{\LARGE \bf
Struct-MDC: Mesh-Refined Unsupervised Depth Completion Leveraging Structural Regularities from Visual SLAM}

\author{
Jinwoo Jeon$^{1}$, 
Hyunjun Lim$^{1}$,
Dong-Uk Seo$^{1}$,~\IEEEmembership{Student~Member,~IEEE}, 
\\and Hyun Myung$^{2*}$,~\IEEEmembership{Senior~Member,~IEEE}
\thanks{This work has been supported by the Unmanned Swarm CPS Research Laboratory program of Defense Acquisition Program Administration and Agency for Defense Development. (No. 2111G5-911257202)}
\thanks{$^{1}$J. Jeon, H. Lim, and D. Seo are with School of Electrical Engineering, Korea Advanced Institute of Science and Technology (KAIST), Daejeon, 34141, Republic of Korea
{\tt\small \{zinuok, tp02134, dongukseo\}@kaist.ac.kr}}%
\thanks{$^{2}$Corresponding author: Prof. Hyun Myung is with School of Electrical Engineering at KAIST, Daejeon, Republic of Korea
{\tt\small hmyung@kaist.ac.kr}}
}

\begin{document}

\maketitle

\begin{abstract}
Feature-based visual simultaneous localization and mapping (SLAM) methods only estimate the depth of \textcolor{color1}{extracted} features, generating a sparse depth map. 
To \textcolor{color1}{solve this} sparsity \textcolor{color1}{problem}, depth completion tasks that estimate a dense depth from a sparse depth have gained \textcolor{color2}{significant importance in robotic applications like exploration.}
Existing methodologies \textcolor{color1}{that use} sparse depth from visual SLAM mainly \textcolor{color1}{employ} point features. However, point features have limitations \textcolor{color1}{in preserving} structural regularities \textcolor{color1}{owing} to texture-less \textcolor{color1}{environments} and sparsity problems. 
To \textcolor{color1}{deal with} these \textcolor{color1}{issues}, we perform depth completion \textcolor{color1}{with} visual SLAM \textcolor{color1}{using line features, which} can better contain structural regularities  \textcolor{color1}{than point features}. 
The proposed methodology creates a convex hull region by performing constrained Delaunay triangulation with depth interpolation using line features. However, the generated depth \textcolor{color1}{includes} low-frequency information and is discontinuous at the convex hull boundary. 
Therefore, we propose a mesh depth refinement \textcolor{color1}{(MDR) module} to \textcolor{color1}{address} this problem. The MDR \textcolor{color1}{module} effectively transfers the high-frequency detail\textcolor{color1}{s} of \textcolor{color1}{an} input image to the interpolated depth and plays a vital role in bridging the conventional and deep learning-based approach\textcolor{color1}{es}.
The Struct-MDC outperforms other state-of-the-art algorithms on public and \textcolor{color1}{our custom} datasets, \textcolor{color1}{and} even \textcolor{color1}{outperforms} supervised methodologies for some metrics. In addition, the effectiveness of the proposed MDR module is verified \textcolor{color1}{by} a rigorous ablation study.
\end{abstract}


\input{1.introduction}
\input{2.relatedworks}
\input{3.proposed}

\input{4.experiments}
\input{5.conclusion}

\bibliographystyle{IEEEtran}
\bibliography{references}

\end{document}

%% file: 1.introduction.tex
\section{Introduction}

\begin{figure}[t]
    \centering
    \includegraphics[width=\linewidth]{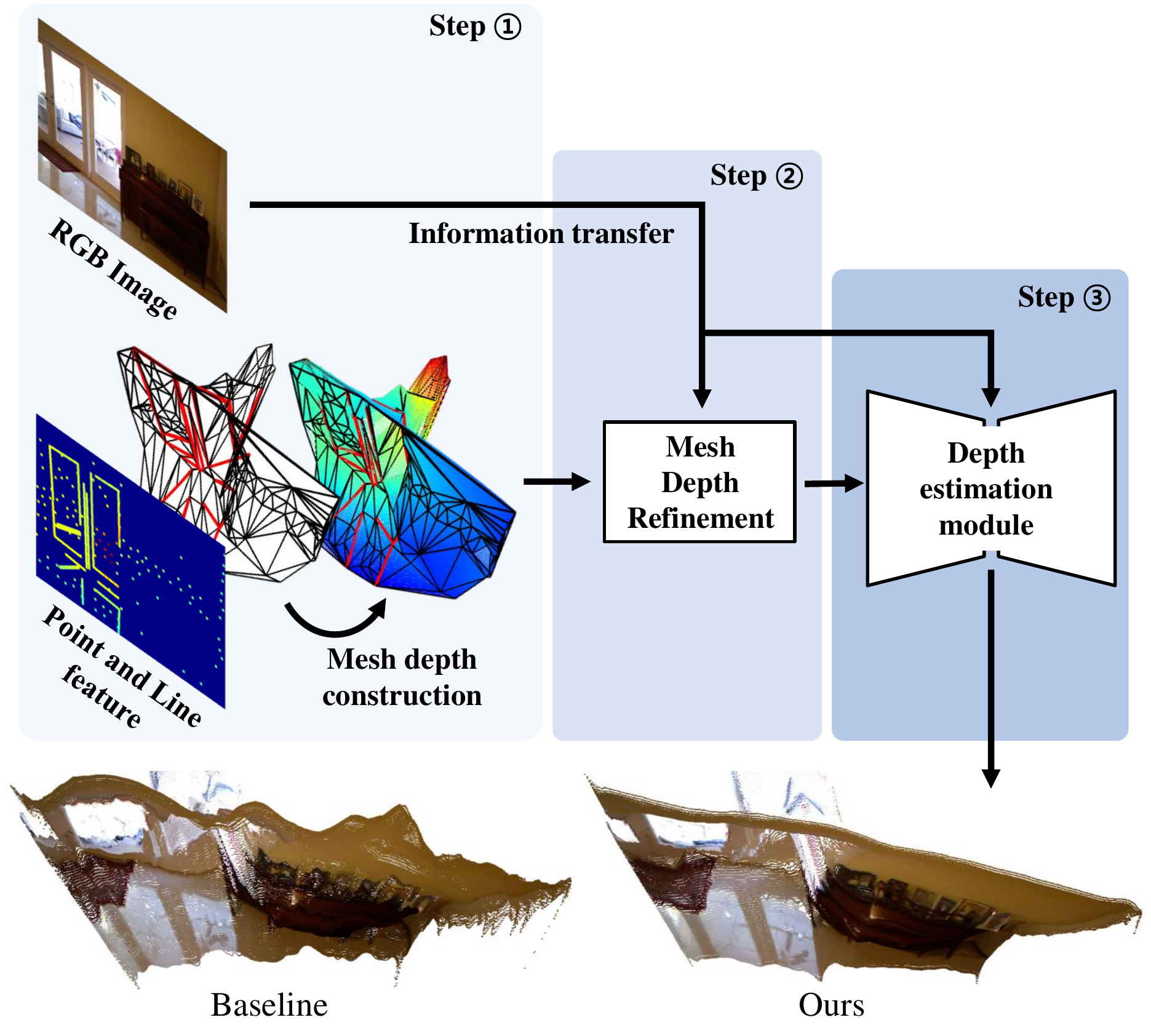}
    \caption{The proposed \textcolor{color1}{three}-step method for depth completion. \textcolor{color2}{KBNet~\cite{baseline} is employed as a baseline in the third step.} Our method effectively capture\textcolor{color1}{s} object boundary using line features, \textcolor{color1}{which are} colored in red, from visual SLAM.}
    \label{fig:cover}
\end{figure}

Feature-based visual simultaneous localization and mapping (SLAM) methods use only partial information of an image, \textcolor{color1}{and are} efficient \textcolor{color1}{for computationally complex cases}. Therefore, compared \textcolor{color1}{with} direct methods\textcolor{color1}{, which} use complete pixel information of \textcolor{color1}{an} image, indirect methods can be \textcolor{color1}{commonly} used on devices with limited computing power \cite{bench2}\textcolor{color1}{,} and many studies \textcolor{color1}{have} been conducted on them \cite{vins, kimera}. 
However, there is no \textcolor{color1}{method} \textcolor{color1}{for determining} the depth information \textcolor{color1}{of the remaining pixels} that \textcolor{color1}{are} not extracted as features \textcolor{color1}{by} the indirect methods. \textcolor{color1}{In addition}, the maximum number of \textcolor{color1}{used} features is limited \textcolor{color1}{to} tens to hundreds per input image frame to \textcolor{color1}{reduce} the computation, \textcolor{color1}{thereby causing sparsity of the depth data}. Therefore, the sparse \textcolor{color1}{depth} map\textcolor{color1}{s} generated \textcolor{color1}{by} indirect methods provide \textcolor{color1}{in}sufficient topology or structural information \textcolor{color1}{of the surroundings of a} robot. These shortcomings make it difficult for indirect methods to be used directly for exploration \textcolor{color1}{and} path planning, which \textcolor{color1}{are} essential applications of SLAM.

To \textcolor{color1}{deal} with the map sparsity problem of indirect methods, depth completion task\textcolor{color1}{s} \textcolor{color1}{are} being actively studied.
\textcolor{color1}{D}epth completion usually estimates a dense depth \textcolor{color1}{using} a sparse depth information \textcolor{color1}{from} visual SLAM with deep learning. \textcolor{color1}{Different from} mono\textcolor{color1}{-}depth prediction \cite{scale_amb2}, which suffer\textcolor{color1}{s} from scale ambiguity, \textcolor{color1}{the} absolute scale from the sparse feature depth facilitates \textcolor{color1}{depth completion} close to the ground truth.
However, most existing depth completion methods rely only on implicit learning of their models to estimate a dense depth. As pointed out in \cite{copypaste1}, if all the inference results solely rely on deep learning, \textcolor{color1}{a} model may fall into a local minim\textcolor{color1}{um} to fill empty spaces with only copy and paste to minimize \textcolor{color1}{the} loss. Therefore, countermeasure\textcolor{color1}{s are} needed to resolve these problems.

Recent studies on depth completion from visual SLAM \textcolor{color1}{have} mainly \textcolor{color1}{focused on the generation of} dense depth \textcolor{color1}{from} point feature depth.
However, in \textcolor{color1}{a} methodology \textcolor{color1}{that utilizes} the point features, it is difficult to \textcolor{color1}{capture} the complete structural regularities of the environment. In addition, point features have \textcolor{color1}{the disadvantage} of being \textcolor{color1}{susceptible} to dynamic illumination and texture\textcolor{color1}{-}less environments.
In terms of SLAM, studies \textcolor{color1}{on using} line features as new measurements are being actively conducted \cite{plsvo, plf, avoiding} to overcome the shortcomings of point features. 
As in the case of state estimation, line features provide powerful spatial \textcolor{color1}{information} in depth completion task\textcolor{color1}{s} and complement the shortcomings of deep learning-based approach\textcolor{color1}{es}, \textcolor{color1}{in which} depth estimation at the boundary of an object becomes ambiguous. 
However, attempts to apply this new measurement to depth completion have not \textcolor{color1}{yet} been actively conducted. 
Therefore, to resolve the problems \textcolor{color1}{of existing methods caused} by point features, \textcolor{color1}{in this study}, line features are utilized for depth completion. To this end, we propose \textcolor{color1}{the} methodology shown in Fig. \ref{fig:cover}, \textcolor{color1}{which utilizes} most of the advantages of line features, \textcolor{color1}{instead of} simply passing point and line features as model input\textcolor{color1}{s}.

The main contributions of this study are as follows:

\begin{itemize}
\item To the best of our knowledge, the Struct-MDC (mesh-refined unsupervised depth completion leveraging structural regularities from visual SLAM) is the \textcolor{color1}{first to} introduce line features from line-based SLAM into a depth completion task.

\item \textcolor{color1}{To} utilize line feature\textcolor{color1}{s} efficiently, unlike other existing \textcolor{color1}{deep learning-based} methodologies that rely on implicit learning, we propose a mesh depth refinement \textcolor{color1}{(MDR) module} with an explicit \textcolor{color1}{three}-step approach: structural sketch, refinement, and estimation with constrained Delaunay triangulation \textcolor{color1}{(CDT)} \cite{cdt} using line features.

\item To prove the validity of Struct-MDC, \textcolor{color1}{the} state-of-the-art performance for depth completion tasks is verified \textcolor{color1}{by} experiments on public datasets and our custom dataset. The code and dataset of the proposed algorithm are publicly available at: \url{https://github.com/zinuok/line_depth_completion}.
\end{itemize}

The \textcolor{color1}{remainder} of \textcolor{color1}{this} paper is organized as follows: Section \ref{sec:related} reviews related \textcolor{color1}{studies}. Section \ref{sec:proposed} describes the proposed method in detail and Section \ref{sec:exp_result} \textcolor{color1}{presents a detailed analysis of} the experimental results. Finally, Section \ref{sec:cons} summarizes our contributions \textcolor{color1}{as well as} future works.

%% file: 2.relatedworks.tex

\section{Related Works} \label{sec:related}

\subsection{Supervised Depth Completion}

In \textcolor{color1}{a} supervised method, the model learns to directly minimize the error between \textcolor{color1}{a} ground truth depth and its estimation.
The RGB and sparse depth \textcolor{color1}{images} have different modalities. 
\textcolor{color1}{To fuse} the heterogeneous  modality inputs, 
early fusion \cite{s2d},
late fusion \cite{late2}, 
and multi-stage fusion \cite{multi_hms} \textcolor{color1}{methods} have been proposed. 
\textcolor{color1}{Using uncertainty estimation,} Teixeira \textit{et al.} \cite{aerial_view} effectively dealt with depth completion on an aerial platform \textcolor{color1}{involving diverse} depth and viewpoint\textcolor{color1}{s}.
Sartipi \textit{et al.} \cite{detph_from_vislam} utilized surface normal prediction and plane detection for depth enrichment of sparse input\textcolor{color1}{s}, focusing on \textcolor{color1}{a} planar structure that is \textcolor{color1}{frequently} observed in indoor environments.
Matsuki \textit{et al.} \cite{codemap} combined prediction from a variational \textcolor{color1}{autoencoder} network with conventional multiple view optimization.
Supervised methods facilitate easy learning\textcolor{color1}{,} with \textcolor{color1}{excellent} performance. However, there is a limitation \textcolor{color1}{in} that accurately \textcolor{color1}{acquiring} \textcolor{color1}{numerous} ground truth depth in the real world \textcolor{color1}{is difficult}, promoting the need for unsupervised learning.

\subsection{Unsupervised Depth Completion}

In unsupervised depth estimation framework\textcolor{color1}{s}, \textcolor{color1}{the} photometric discrepancy between actual and reconstructed image\textcolor{color1}{s} using \textcolor{color1}{a} multiple view geometry has been used as a supervisory signal \cite{uns_phot2, uns_phot3}.
This scheme \textcolor{color1}{is also applicable} to unsupervised depth completion frameworks \cite{ss-s2d, selfdeco}.
\textcolor{color1}{Owing} to the lack of ground truth \textcolor{color1}{depth}, several techniques are used to assist model learning in unsupervised methodolog\textcolor{color1}{ies}. 
Among them, classical techniques may help \textcolor{color1}{a} model understand the scene topology. 
In \cite{void}, \textcolor{color1}{a} lightweight network using Delaunay triangulation was presented. However, \textcolor{color1}{its} accuracy is limited because the triangulation generated by point features does not fit well \textcolor{color1}{in the case of} complex structure\textcolor{color1}{s}. 
Wong \textit{et al.} \cite{baseline} utilized \textcolor{color1}{the} camera's intrinsic parameter as \textcolor{color1}{an} input to improve the generalization performance of \textcolor{color1}{a} model \textcolor{color1}{by} feature encoding back-project\textcolor{color1}{ion} into 3D space.
However, \textcolor{color1}{this method} still \textcolor{color1}{depends} on \textcolor{color1}{a} front-end pooling layer to densify sparse point depth. 
In \cite{learning_top}, \textcolor{color1}{a} topology prior learned from \textcolor{color1}{a} synthetic dataset was refined on the real-world dataset. However, max pooling used to fill the empty depth misses the detailed depth information of \textcolor{color1}{an} object.
The aforementioned methods still have problems resulting from point features. 
To \textcolor{color1}{overcome} this \textcolor{color1}{issue}, \textcolor{color1}{there is a} need to introduce a new \textcolor{color1}{type} of measurement.

\subsection{Edge \textcolor{color1}{A}wareness and Depth Refinement}

Edges play vital role\textcolor{color1}{s} in depth estimation \textcolor{color1}{because they} can \textcolor{color1}{identify the} depth discontinuity at \textcolor{color1}{an} object boundary. 
\textcolor{color1}{Motivated} by \cite{ransac}, Jiang \textit{et al.} \cite{plnet} adopted \textcolor{color1}{a} random point selection scheme to utilize \textcolor{color1}{pervasive} line and plane structures in an indoor environment as supervisory signals. 
In \cite{vplnet}, the Manhattan world assumption and vanishing point clustering were employed to estimate surface normals \textcolor{color1}{with increased accuracy}.
Ferstl \textit{et al.} \cite{edge_superes} investigated the use of the super-resolution methodology of depth image\textcolor{color1}{s} \textcolor{color1}{by} the edge-aware energy optimization.
\textcolor{color1}{E}dge awareness can also support \textcolor{color1}{a} depth refinement task \cite{ddrnet}\textcolor{color1}{, which} refine\textcolor{color1}{s} coarse depth estimated by \textcolor{color1}{a} model guided by an image. 
Khamsis \textit{et al.} \cite{stereonet} added detailed information to \textcolor{color1}{a} disparity map estimated from stereo images by dilating or eroding around the edges with iterative hierarchical refinement.
Qi \textit{et al.} \cite{geonet++} refined initial surface normal and depth estimation\textcolor{color1}{s} by generating a weight map with edges extracted from \textcolor{color1}{an} image. 
\textcolor{color1}{Because} depth completion inherits the essential properties of depth estimation, it can benefit from \textcolor{color1}{edge awareness}. However, attempts to introduce the \textcolor{color1}{edge awareness} advantage from line feature depth into \textcolor{color1}{a} depth completion framework have not been actively \textcolor{color1}{conducted}.

%% file: 3.proposed.tex
\section{Struct-MDC} \label{sec:proposed}

\subsection{Overall Framework}\label{sec:overall_framework}

\begin{figure}[thb!]
    \centering
    \includegraphics[width=\columnwidth]{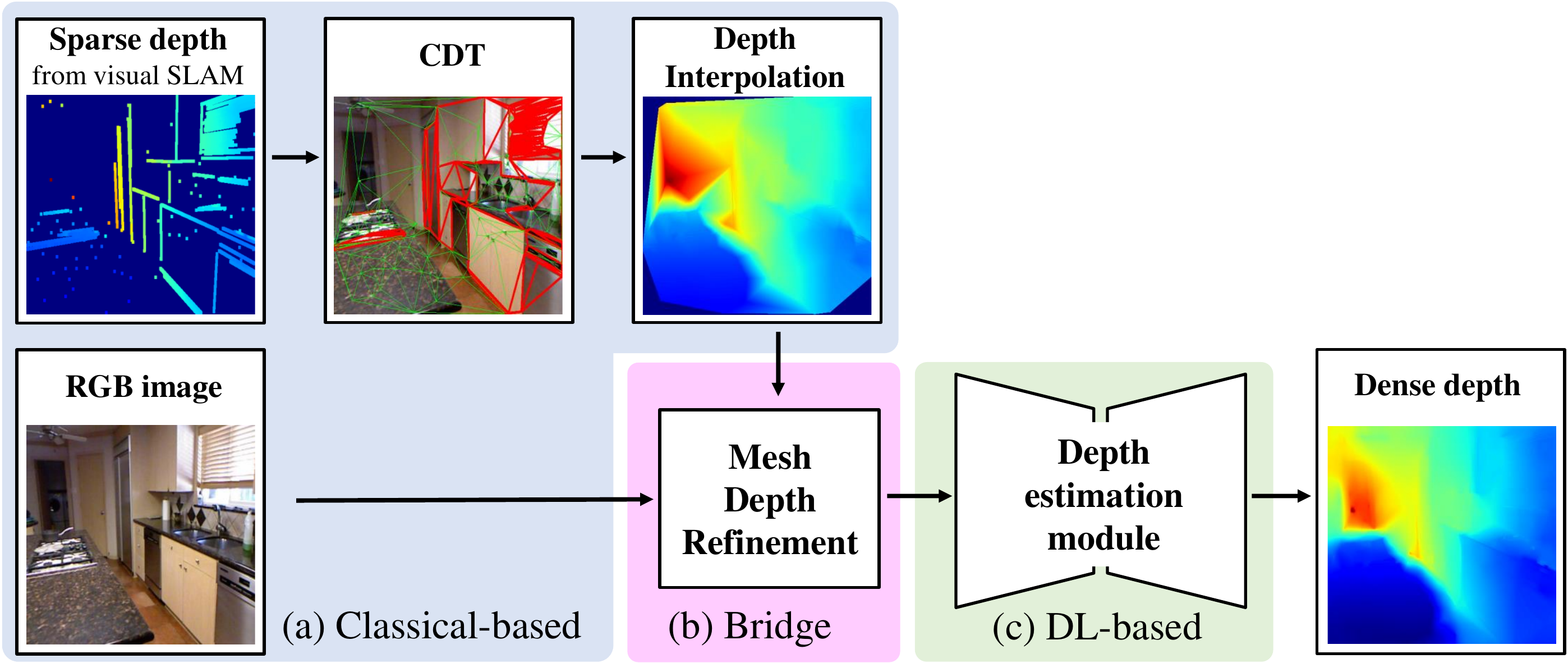}
    \caption{Overall network architecture of struct-MDC: dense depth is estimated through 
        (a) structural sketch and interpolation, 
        (b) refinement, and
        (c) estimation.} 
    \label{fig:overall_arch}
    \vspace{-0.45cm}
\end{figure} 

Our \textcolor{color1}{objective} is to estimate \textcolor{color1}{a} pixel-wise depth, given an RGB image and \textcolor{color1}{a} sparse feature depth from visual SLAM. The overall framework of struct-MDC is shown in Fig.~\ref{fig:overall_arch}. The proposed method \textcolor{color1}{is} explicitly \textcolor{color1}{split} into three steps: structural sketch, refinement, and estimation. 
In the structural sketch step, struct-MDC uses point and line features extracted from visual SLAM. Using \textcolor{color1}{these} features, \textcolor{color1}{a} mesh with structural regularities is generated \textcolor{color1}{by} CDT. 
Using the constructed mesh, \textcolor{color1}{a} rough mesh depth is generated \textcolor{color1}{by} depth interpolation.  
\textcolor{color1}{Owing} to discontinuit\textcolor{color1}{ies} and low-frequency detail problems, the generated mesh depth is \textcolor{color1}{un}suitable \textcolor{color1}{for direct use} as an input \textcolor{color1}{to} the main module. Therefore, \textcolor{color1}{an} \textcolor{color1}{MDR} module is proposed. The MDR module acts as a bridge to generate \textcolor{color1}{a} refined mesh depth and \textcolor{color1}{transfer} detailed information from the \textcolor{color1}{given} RGB image to the rough mesh depth.
At last, the final dense depth is estimated \textcolor{color1}{by} the main module. As a baseline, the main encoder-decoder architecture of \cite{baseline} is adopted. Its front-end sparse-to-dense module is excluded because our structural sketch and interpolation followed by refinement can effectively replace the module. 
We assume that the surrounding environment structure is \textcolor{color1}{piecewise} planar and rectilinear for the framework \textcolor{color1}{proposed} in this \textcolor{color1}{study}.
The struct-MDC is \textcolor{color1}{expressed} as follows:
\begin{equation}
\begin{gathered}
\mathbf{z}_d = f \circ g \circ h (I, \mathcal{P}, \mathcal{L}, \mathbf{K}),
\\
\mathcal{P} := \bigcup^m_{i=1}(\mathbf{x}_{p_i}, z_{p_i}), 
\\
\mathcal{L} := \bigcup^n_{j=1}(l_j(\mathbf{x}_{s_j}; \mathbf{x}_{e_j}), z_{s_j}, z_{e_j}),
\end{gathered}
\end{equation}
where $\mathbf{z}_d$ denotes the densified depth; $f$, $g$, and $h$ denote functions of the depth estimation, refinement, and structural sketch \textcolor{color1}{steps}; 
$I$ and $\mathbf{K}$ \textcolor{color1}{denote} the RGB image and the camera's intrinsic parameter; 
$\mathcal{P}$ and $\mathcal{L}$ denote sets of point and line features with depth; 
$\mathbf{x}_{p_i}$, $\mathbf{x}_{s_j}$, and $\mathbf{x}_{e_j}$ represent \textcolor{color1}{the} $i$-th 2D point feature, \textcolor{color1}{the start and end points of the} $j$-th 2D line feature $l_j$; and $m$ and $n$ are the numbers of point and line features; $z_{p_i}$, $z_{s_j}$, and $z_{e_j}$ denote the depth of the $i$-th point feature,  those of the start and end points of the $j$-th line feature, respectively.


\subsection{Structural \textcolor{color1}{S}ketch \textcolor{color1}{by} Constrained Delaunay Triangulation}\label{sec:feature_extraction}

\begin{figure}[t!]
    \centering
    \subfigure[]{\includegraphics[width=0.32\linewidth]{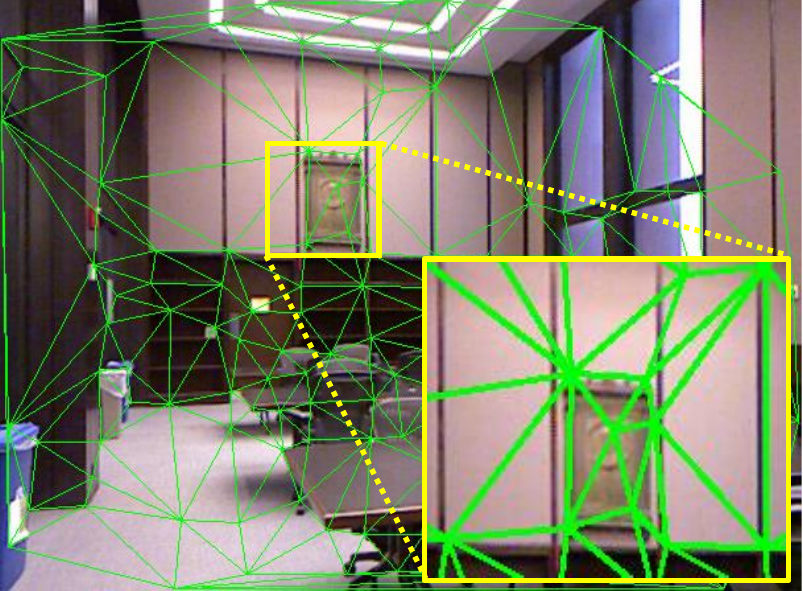}\label{fig:DT}}
    \subfigure[]{\includegraphics[width=0.32\linewidth]{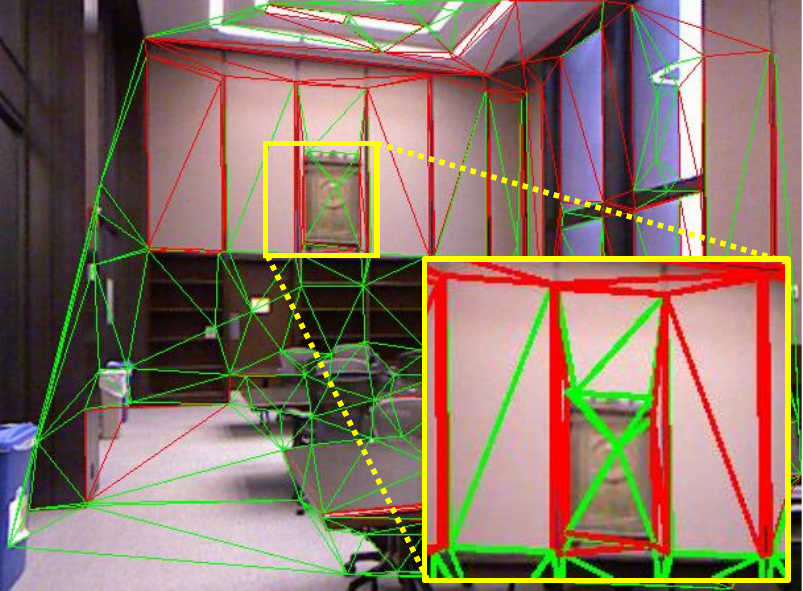}\label{fig:CDT}}
    \subfigure[]{\includegraphics[width=0.32\linewidth]{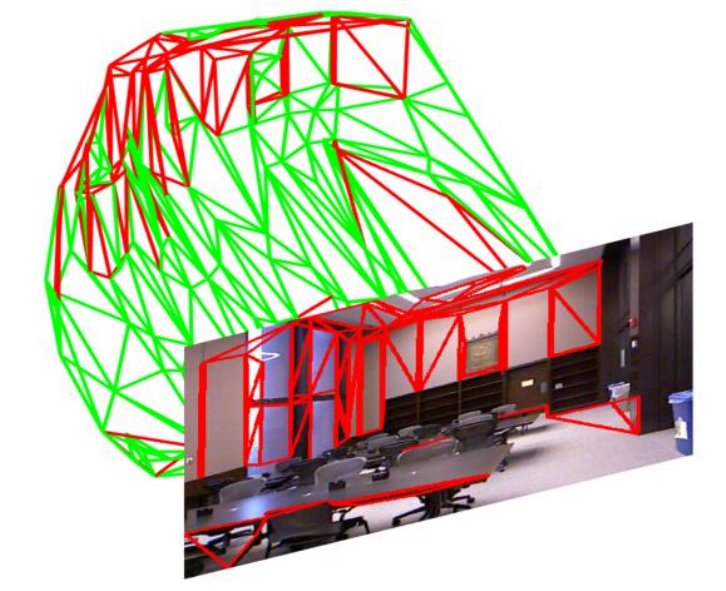}\label{fig:3d_mesh}}
    \caption{Constructed mesh\textcolor{color1}{es} with \subref{fig:DT} only point features and \subref{fig:CDT} point and line features (ours). \subref{fig:3d_mesh} Back-projected mesh into 3D space. The line features are colored in red.}
    \label{fig:mesh}
    \vspace{-0.3cm}
\end{figure}

\begin{figure}[t!]
    \centering
    \subfigure[]{\label{fig:before_interp}\includegraphics[width=0.45\linewidth,height=0.24\linewidth]{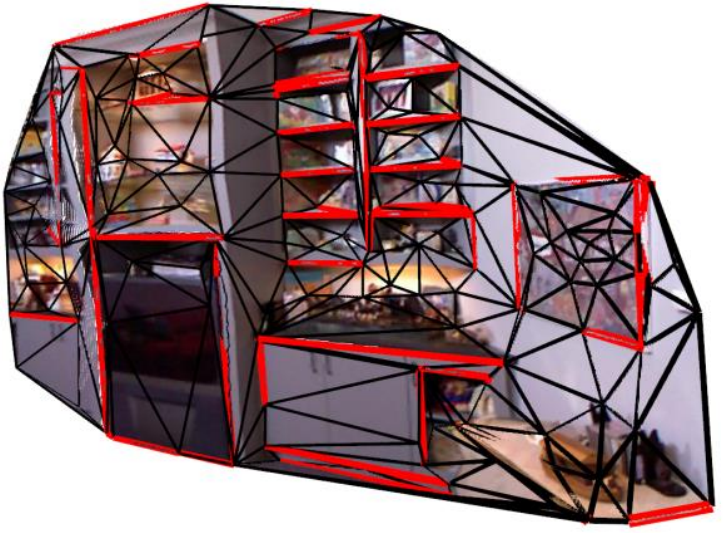}}
    \subfigure[]{\label{fig:after_interp}\includegraphics[width=0.49\linewidth,height=0.25\linewidth]{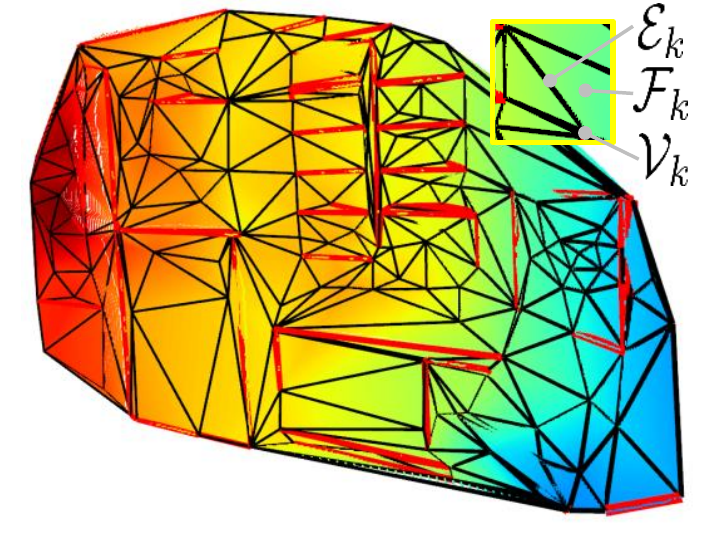}}
    \caption{Before \subref{fig:before_interp} and after \subref{fig:after_interp} \textcolor{color1}{mesh depth interpolation}. The line features are colored in red.}
    \label{fig:mesh_interp}
    \vspace{-0.3cm}
\end{figure}

In this stage, \textcolor{color1}{a} rough mesh depth is generated from \textcolor{color1}{a} sparse feature depth based on the classical methodology. \textcolor{color1}{Construction of the} rough mesh depth \textcolor{color1}{requires} mesh generation using features.
As shown in Figs.~\ref{fig:mesh} and \ref{fig:mesh_interp}, line features are formed at the object boundary, supporting our rectilinear assumption. \textcolor{color1}{L}ine features better represent the structural regularities of \textcolor{color1}{an} environment than point features. Therefore, we use CDT to create \textcolor{color1}{a} mesh using point and line features. 
\textcolor{color1}{Different from} \cite{void}\textcolor{color1}{, in} which triangulation \textcolor{color1}{was performed} with only point features, CDT can \textcolor{color1}{strongly} leverage the structural regularities of a human-made environment\textcolor{color1}{, because the} detected line features are included as constrained edges\textcolor{color1}{,} as shown in Figs.~\ref{fig:mesh}\subref{fig:DT} and \subref{fig:CDT}. \textcolor{color1}{Owing to} this advantage, CDT is performed using point and line features as follows: 
\begin{equation}
\begin{gathered}
(\mathcal{F}, \mathcal{V}, \mathcal{E}) 
= \mathcal{C} \left(\bigcup^m_{i=1}\mathbf{x}_{p_i}, \bigcup^n_{j=1}l_j(\mathbf{x}_{s_j}; \mathbf{x}_{e_j})\right),
\label{eq:cdt}
\end{gathered}
\end{equation}
where $\mathcal{C}$ denotes a CDT generation function that uses point and line features as input; $\mathcal{F}$, $\mathcal{V}$, and $\mathcal{E}$ \textcolor{color1}{denote} triangle facets, vertices, and edges of CDT, respectively. Using CDT, \textcolor{color1}{a} line feature is explicitly included in the $\mathcal{E}$ as constrained edges.


Using the depth information of \textcolor{color1}{the} point and line features included in the $\mathcal{V}$ and $\mathcal{E}$, the 2D mesh on the image is back-projected into a 3D space\textcolor{color1}{,} as shown in Fig.~\ref{fig:mesh}\subref{fig:3d_mesh}.
The 3D mesh \textcolor{color1}{contains the} depth information only at \textcolor{color1}{the} vertices and edges corresponding to the point and line features. Therefore, \textcolor{color1}{to fill} the depth of triangle facet using the boundary depth information, linear interpolation is performed\textcolor{color1}{,} as shown in Fig. \ref{fig:mesh_interp}\textcolor{color1}{,} and \textcolor{color1}{it is expressed} as follows: 
\begin{equation}
\begin{gathered}
\mathbf{z}_m 
= h(\mathcal{F}, \mathcal{V}, \mathcal{E}, \mathcal{P}, \mathcal{L}),
\label{eq:f_interp}
\end{gathered}
\end{equation}
where $h$ \textcolor{color1}{denotes} back-projection into a 3D space followed by linear depth interpolation; $\mathbf{z}_m$ represents the rough mesh depth.



\subsection{Mesh Depth Refinement Module}\label{sec:mdr_module}

\begin{figure}[t!]
    \centering
    \includegraphics[width=\columnwidth, height=3.8cm]{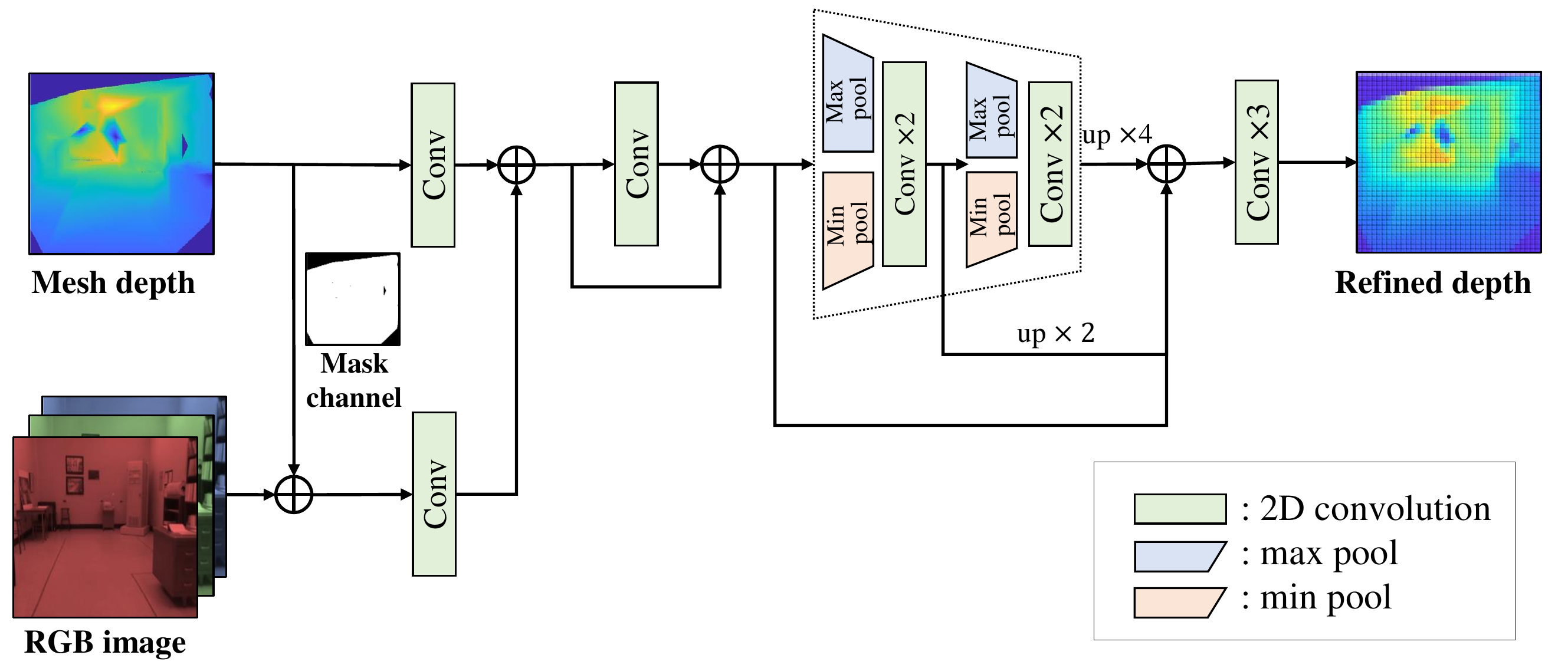}
    \caption{Proposed \textcolor{color1}{m}esh \textcolor{color1}{d}epth \textcolor{color1}{r}efinement (MDR) module. Introducing mask channel and parallel pooling facilitates the module to learn the discontinuity and complementary spatial information, respectively.} 
    \label{fig:mdr}
    \vspace{-0.3cm}
\end{figure}

There are two major problems with the generated mesh depth\textcolor{color1}{: depth discontinuity and low-frequency depth.} 
The point and line feature sets are bounded inside the image plane. 
At \textcolor{color1}{this time}, \textcolor{color1}{because} the mesh constructed \textcolor{color1}{by} CDT forms a convex hull, \textcolor{color1}{its} area cannot \textcolor{color1}{exceed} the entire image \textcolor{color1}{area}. Therefore, \textcolor{color1}{a} depth outside the convex hull cannot be interpolated, \textcolor{color1}{producing a} depth discontinuity at its boundary. The area padded with zeros adversely affects the neuron activation of the network.
Another major problem is that \textcolor{color1}{a simple} linear interpolation will eliminate high-frequency information in the interpolated depth. Therefore, detail\textcolor{color1}{s} of the actual object \textcolor{color1}{are} lost. 
To solve these problems, the MDR module \textcolor{color1}{shown in} Fig.~\ref{fig:mdr} is proposed. 
\textcolor{color1}{Motivated} by \cite{ddrnet}, the MDR \textcolor{color1}{module} performs refinement by transferring high-frequency detail from an RGB image to low-frequency discontinuous mesh depth as follows: 
\begin{align}
    \mathbf{z}_r = g(I, \mathbf{z}_m),
\end{align}
where $\mathbf{z}_r$ denotes the refined mesh depth.
Specifically, to deal with the depth discontinuity problem, the binary mask channel extracted from the mesh depth is concatenated in the form of \textcolor{color1}{hypercolumns}. The mask channel transmits information \textcolor{color1}{about the} discontinuity in advance, enabling the first convolution of the RGB image branch to adaptively transfer information depending on the presence of depth. The concatenated channel also helps deliver information outside the convex hull. 
\textcolor{color1}{Subsequently}, similar to the original module which uses max pooling, \textcolor{color1}{multiscale} feature fusion is performed to learn \textcolor{color1}{the} spatial information. 
In particular, parallel pooling is proposed \textcolor{color1}{for downsampling}. Existing methodologies with only max pooling defect detailed information \textcolor{color1}{because} only the strong activation signals from color and depth images are extracted. \textcolor{color1}{To compensate this deficiency}, min and max pooling are \textcolor{color1}{fused} complementarily as parallel branches.


\subsection{Depth Estimation}\label{sec:depth_est}

For estimating the dense depth, the calibrated back-projection network proposed in \cite{baseline} is adopted as a main module. Their sparse-to-dense module is excluded \textcolor{color1}{because} the proposed structural sketch and refinement steps effectively replace the front-end module. Taking the refined mesh depth as \textcolor{color1}{the} input, the main module estimates \textcolor{color1}{a} dense depth as follows:
\begin{align}
    \mathbf{z}_d = f(I, \mathbf{z}_r, \mathbf{K}),
\end{align}
where $\mathbf{z}_d$ denotes the estimated dense depth. 
In the calibrated back-projection layer, the model \textcolor{color1}{uses} a camera's intrinsic parameter as input. After the encoded feature map from \textcolor{color1}{the} RGB image is back-projected with the encoded depth and the camera's intrinsic parameter, \textcolor{color1}{3D positional encoding is performed}.


\subsection{Loss Function}\label{sec:total_loss}

\textcolor{color1}{The} loss commonly employed for unsupervised depth completion is used \textcolor{color1}{for the regression of the proposed network}. Please note that the proposed MDR module can be integrated into the entire system without any additional loss function. Therefore, some notations used in the baseline \cite{baseline} are \textcolor{color1}{adopted in} this \textcolor{color1}{study}. The overall loss is \textcolor{color1}{calculated} as follows: 
\begin{align}
L = w_p l_p + w_s l_s + w_l l_l,
\end{align}
where $l_p$, $l_s$, and $l_l$ denote the photometric consistency, sparse depth consistency, and local smoothness loss; $w_p$, $w_s$, and $w_l$ denote their corresponding weights, respectively.

\subsubsection{Photometric consistency loss}
Using the estimated depth and \textcolor{color1}{the} relative pose between consecutive frames, the photometric error between the reconstructed and actual image\textcolor{color1}{s} can be defined as follows: 
\begin{equation}    
\begin{aligned}
l_p = \frac{1}{|\Omega|} 
\sum_{t' \in \Tau}^{}
\sum_{p \in \Omega}^{} 
w_1 \Vert I(\mathbf{x}) - I_{t' \rightarrow t}(\mathbf{x}) \Vert 
\\
+
w_2 (1 - {SSIM}(I(\mathbf{x}) - I_{t' \rightarrow t}(\mathbf{x}))),
\end{aligned}
\end{equation}
where $\Tau$ denotes a time sequence; $I_{t' \rightarrow t}$ and $I_t$ denote \textcolor{color1}{the} warped image from $t'$ to $t$ and \textcolor{color1}{the} actual image at $t$; $\mathbf{x}$ and $\Omega$ represent \textcolor{color1}{a} single pixel and \textcolor{color1}{the} collection of all pixels of the image; and $w_1$ and $w_2$ denote the corresponding photometric discrepancy weight\textcolor{color1}{s}, respectively. 
$SSIM$ is \textcolor{color1}{the} structural similarity index defined in \cite{ssim} as follows: 
\begin{equation}    
\begin{aligned}
SSIM(\alpha,\beta) = \frac
{(2 \mu_\alpha \mu_\beta + C_1)(2 \sigma_{\alpha\beta} + C_2)}
{(\mu^2_\alpha + \mu^2_\beta + C_1)(\sigma^2_\alpha + \sigma^2_\beta + C_2)},
\end{aligned}
\end{equation}
where $\alpha$ and $\beta$ denote images to compare; $\mu$ and $\sigma$ denote the mean intensity and the standard deviation; $C_1$ and $C_2$ represent small constants for mathematical stability, respectively.

\subsubsection{Sparse depth consistency loss}
\textcolor{color1}{For a} provided feature depth, the regression can \textcolor{color1}{be conducted using} the \textcolor{color1}{information obtained} by measuring the residual between the prediction from the network and the feature depth as follows: 
\begin{equation}
\begin{aligned}
l_s = \frac{1}{|\Omega_v|} \sum_{\mathbf{x} \in \Omega_v}^{} 
{(\Vert \mathbf{z}_d(\mathbf{x}) - \mathbf{z}_f(\mathbf{x}) \Vert)},
\end{aligned}
\end{equation}
where $\mathbf{z}_f$ and $\Omega_v$ denote \textcolor{color1}{the} point and line feature depth \textcolor{color1}{and} a set of valid pixels with \textcolor{color1}{a} sparse depth, respectively.

\subsubsection{Local smoothness loss}
To ensure the smooth\textcolor{color1}{ness of the} depth \textcolor{color1}{of a particular} object, the gradient of the estimated depth is used as follows: 
\begin{align}
l_l = \frac{1}{|\Omega|} \sum_{\mathbf{x} \in \Omega}^{} 
{e^{-|\partial_x I(\mathbf{x})|} |\partial_x \mathbf{z}_d(\mathbf{x})|
+
e^{-|\partial_y I(\mathbf{x})|} |\partial_y \mathbf{z}_d(\mathbf{x})|}.
\end{align}
\textcolor{color1}{A} pixel area with a large image gradient is penalized \textcolor{color1}{to accommodate} the depth discontinuity at the object boundary.


%% file: 4.experiments.tex
\section{Experimental Results}\label{sec:exp_result}

\textcolor{color3}{
This section is organized as follows: The effectiveness of our proposed framework is verified in Section \ref{sec:exp_result}.\textit{C}. An ablation study is conducted for the proposed MDR module in Section \ref{sec:exp_result}.\textit{D}.
}

\subsection{Implementation Details}\label{sec:settings}

Our network is trained on a single NVIDIA TITAN V GPU with 12 GB memory. \textcolor{color1}{The batch} size is set \textcolor{color1}{as} 4 for \textcolor{color1}{the} VOID \cite{void}, \textcolor{color1}{and} NYUv2 \cite{nyuv2} \textcolor{color1}{datasets}, and 2 for our custom \textcolor{color1}{point-line feature and depth (PLAD)} dataset considering the memory capacity. 
For training, the Adam optimizer with exponential decay rates of $\beta_1 = 0.9$ and $\beta_2 = 0.999$ is used.
The learning rate initialized \textcolor{color1}{as} $1 \times 10^{-4}$ decays by half at 20 epochs\textcolor{color1}{,} and is set \textcolor{color1}{as} $5 \times 10^{-5}$ up to 30 epochs.
\textcolor{color1}{In addition}, \textcolor{color1}{the} loss weights are set \textcolor{color1}{as} $w_p=1.0$, $w_s=0.6$, $w_l=0.04$, $w_1=0.15$, and $w_2=0.95$. 
After image undistortion for line extraction and cropping, $608 \times 448$ \textcolor{color1}{size images} are used for \textcolor{color1}{the} VOID and NYUv2 \textcolor{color1}{datasets}, and $1280 \times 720$ size image\textcolor{color1}{s are} used for \textcolor{color1}{the} PLAD \textcolor{color1}{dataset}.

\begin{figure}[t!]
    \centering
    \subfigure[]{\includegraphics[width=0.234\linewidth]{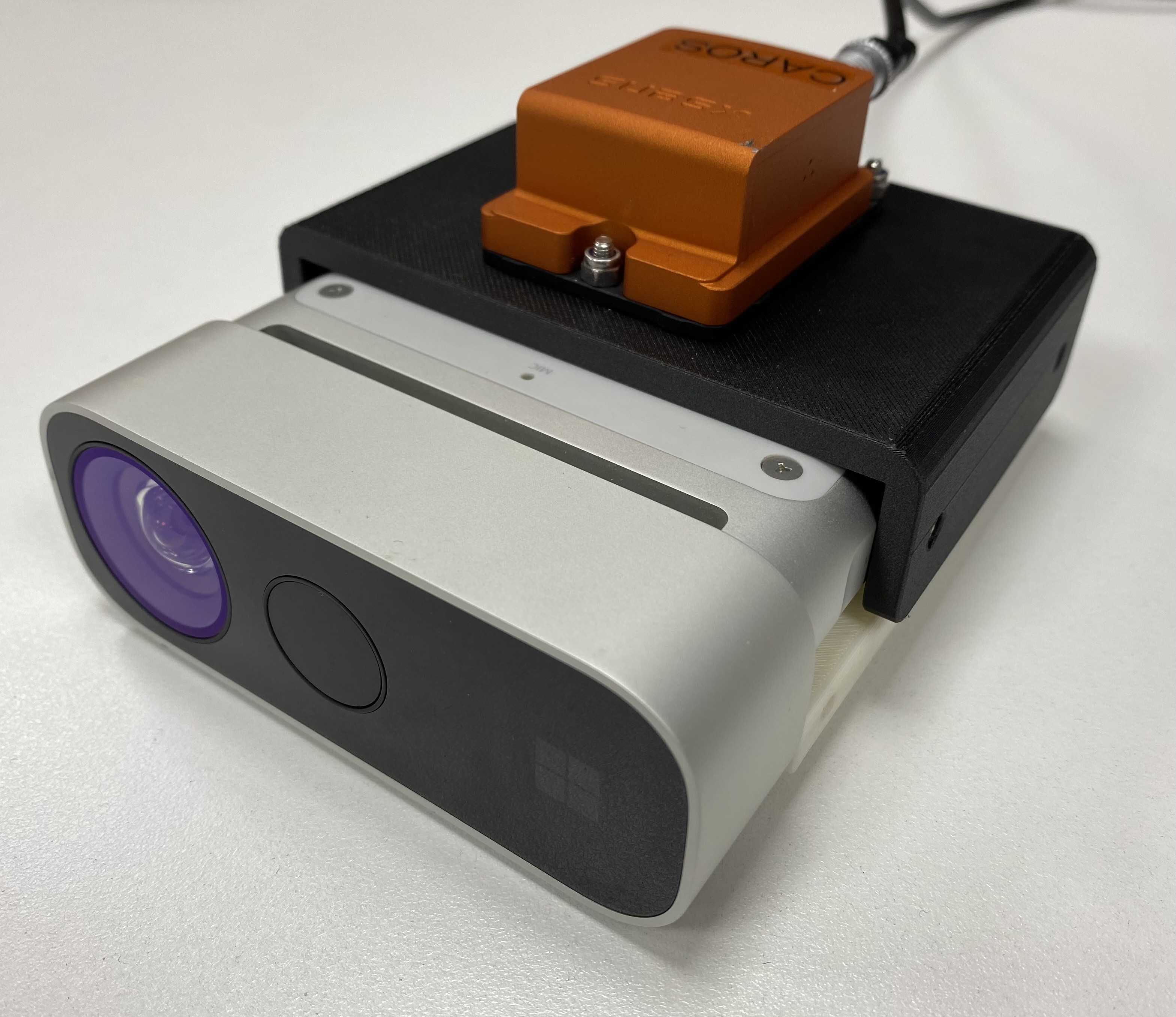}\label{fig:plad_sensor}}
    \subfigure[]{\includegraphics[width=0.365\linewidth]{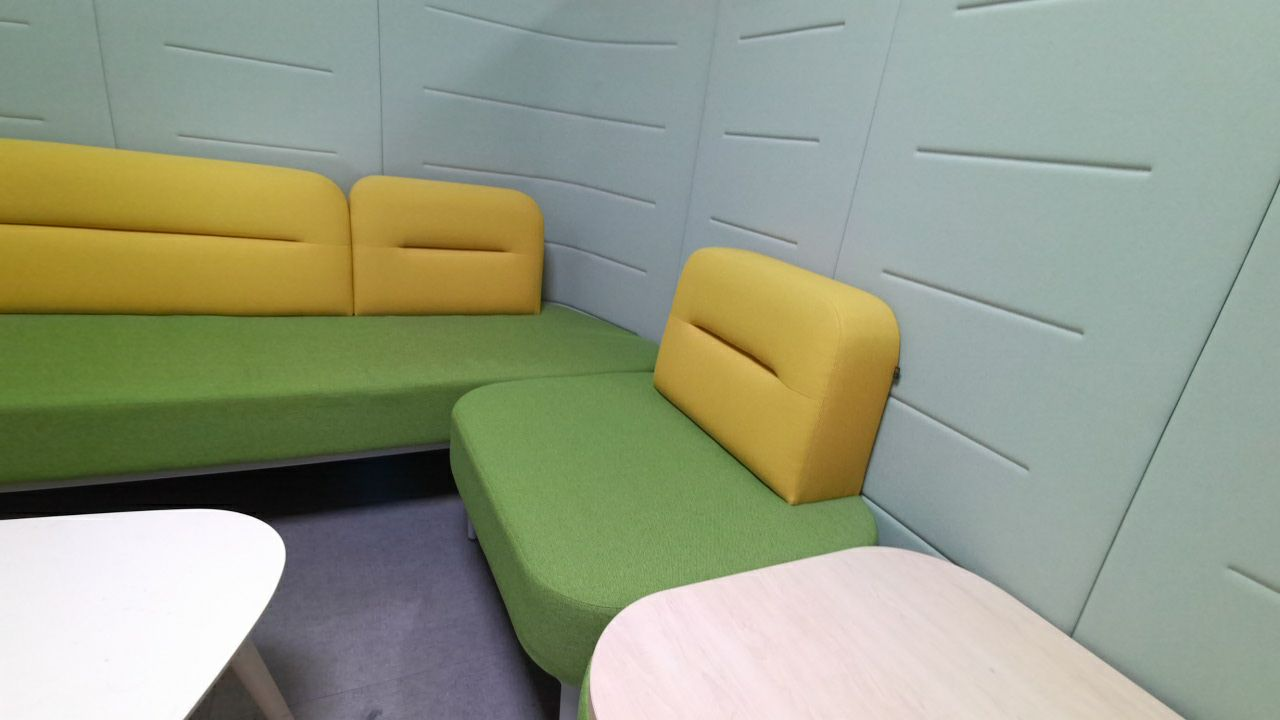}\label{fig:plad_indoor}}
    \subfigure[]{\includegraphics[width=0.365\linewidth]{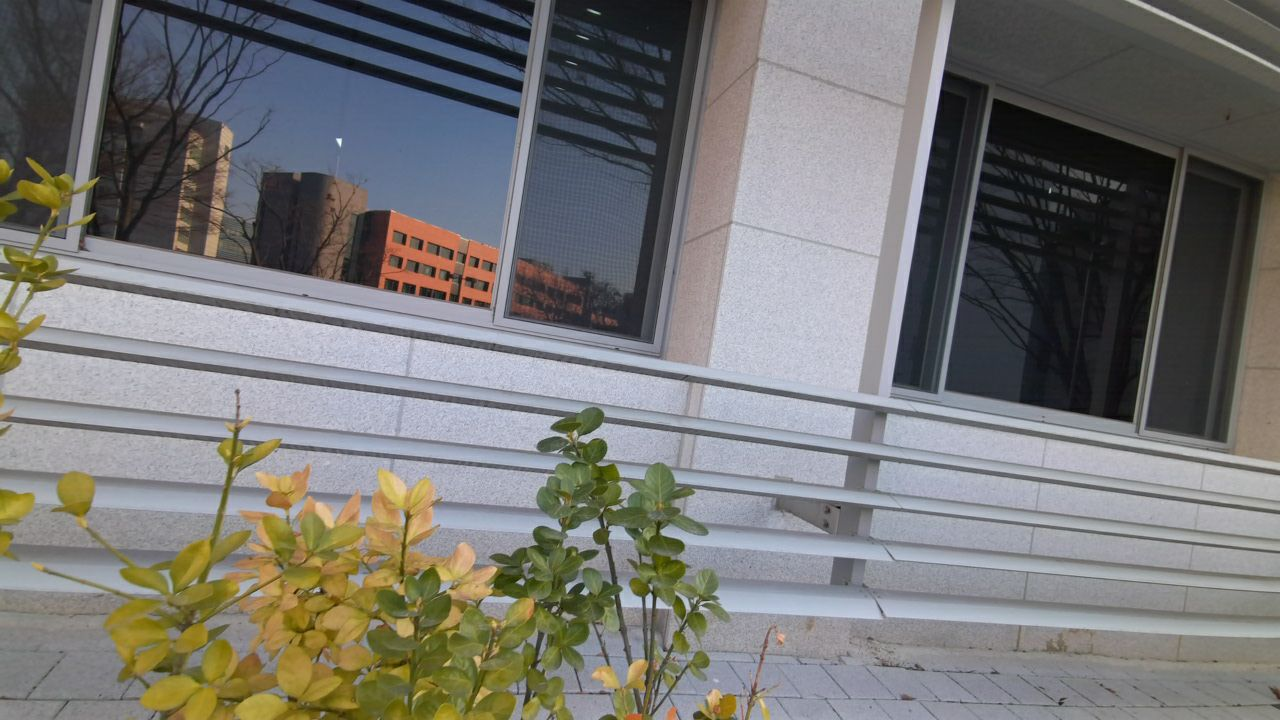}\label{fig:plad_outdoor}}
    \caption{Our PLAD dataset: 
        \subref{fig:plad_sensor} Sensor setup consists of RGB-D camera and IMU. 
        \textcolor{color1}{S}equences \textcolor{color1}{acquired} 
        \subref{fig:plad_indoor} indoor and 
        \subref{fig:plad_outdoor} outdoor
    }
    \label{fig:plad}
\end{figure}

\subsection{\textcolor{color2}{Datasets}}
\subsubsection{VOID}
This dataset provides synchronized RGB images ($640 \times 480$), ground truth depth images ($640 \times 480$), inertial measurement unit (IMU) data from Intel D435i, and sparse point features from visual-inertial odometry \cite{xivo}. 
\textcolor{color1}{Because} Struct-MDC additionally utilizes line features, UV-SLAM \cite{uvslsam} is executed to detect point and line features \textcolor{color1}{and} generate sparse depth. However, \textcolor{color1}{owing} to the challeng\textcolor{color1}{es} with substantial motion blur of the provided sequences, \textcolor{color1}{the} ground truth depth is sampled using the feature matching results of the SLAM front-end.
Of the total 56 sequences, 47 sequences are used for training, \textcolor{color1}{8 for testing, and 1} with poor feature detection and mesh results is excluded.

\subsubsection{NYUv2}
This dataset provides synchronized RGB images ($640 \times 480$) and ground truth depth images ($640 \times 480$) from Microsoft Kinect. 
\textcolor{color1}{Because} this dataset does not provide the IMU data required to execute UV-SLAM, the ground truth depth is sampled as in the VOID dataset.
Post-processed data with filled depth values \textcolor{color1}{are} used as in \cite{s2d}. According to the official split, 48K frames are used for training \textcolor{color1}{and 654 frames} for testing.

\subsubsection{\textcolor{color1}{PLAD}}
\textcolor{color1}{A} sparse depth is inevitably sampled from the ground truth depth in the aforementioned datasets, which \textcolor{color1}{are} far from \textcolor{color1}{the} genuine depth completion from visual SLAM. Therefore, we \textcolor{color1}{generate} a PLAD dataset where sparse depth is provided by line-based visual SLAM to verify Struct-MDC.
As shown in Fig.~\ref{fig:plad}\subref{fig:plad_sensor}, the data \textcolor{color1}{are} acquired with a sensor setup consisting of \textcolor{color1}{an} Azure Kinect for ground truth depth with RGB images ($1280 \times 720$, 15 Hz) and \textcolor{color1}{an} Xsens Mti-100 for IMU data (6-axis, 200 Hz).
The depth and RGB images are synchronized with the feature depth estimated \textcolor{color1}{using} UV-SLAM. Each sequence is acquired in various environments, as shown in Figs.~\ref{fig:plad}\subref{fig:plad_indoor} and \subref{fig:plad_outdoor}.
The PLAD \textcolor{color1}{dataset} consists of 38 sequences, with 34 sequences used for training and 4 sequences for testing. More detailed information \textcolor{color1}{such as the} sensor calibration data is available at: \url{https://github.com/zinuok/line_depth_completion}.

\subsection{Results \textcolor{color1}{of} Analysis}\label{sec:exp_main}
\begin{figure}[t]
    \centering
    \subfigure[]{\includegraphics[width=0.155\linewidth]{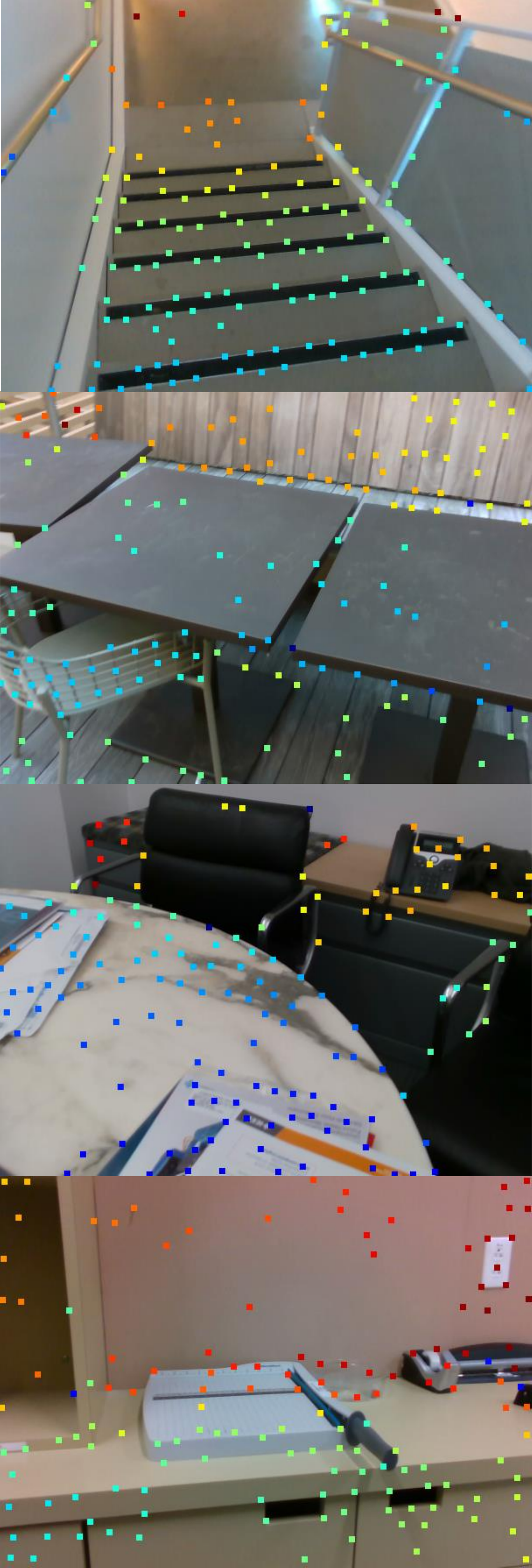}\label{fig:void_point}}
    \subfigure[]{\includegraphics[width=0.155\linewidth]{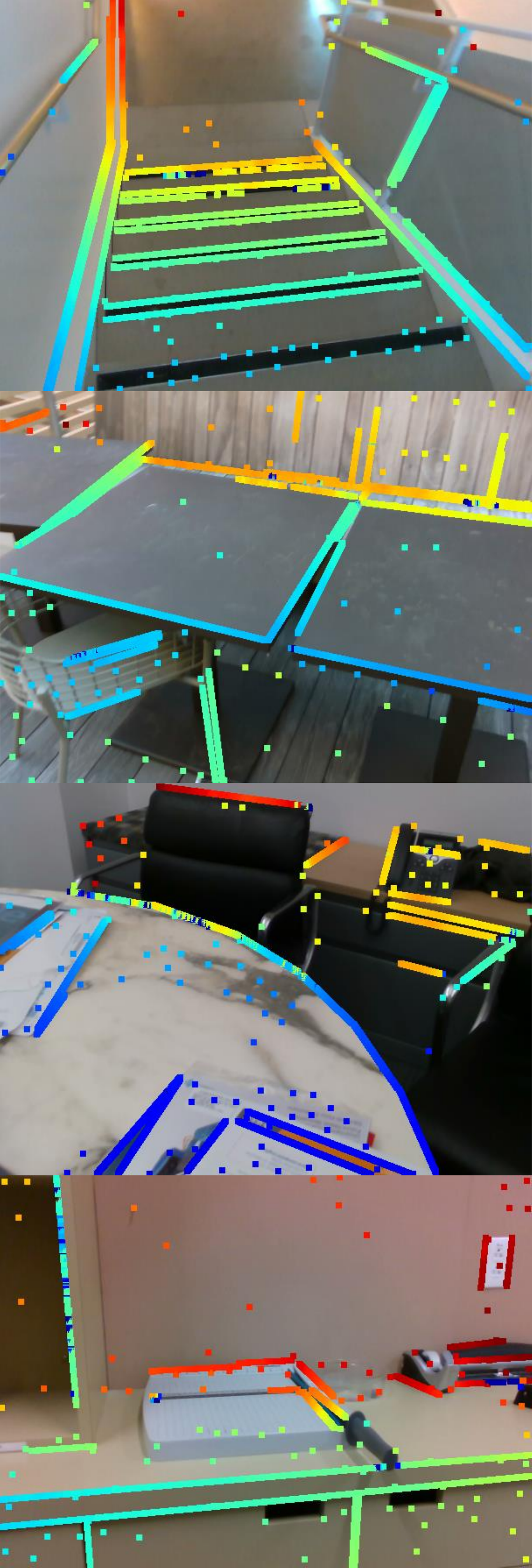}\label{fig:void_line}}
    \subfigure[]{\includegraphics[width=0.155\linewidth]{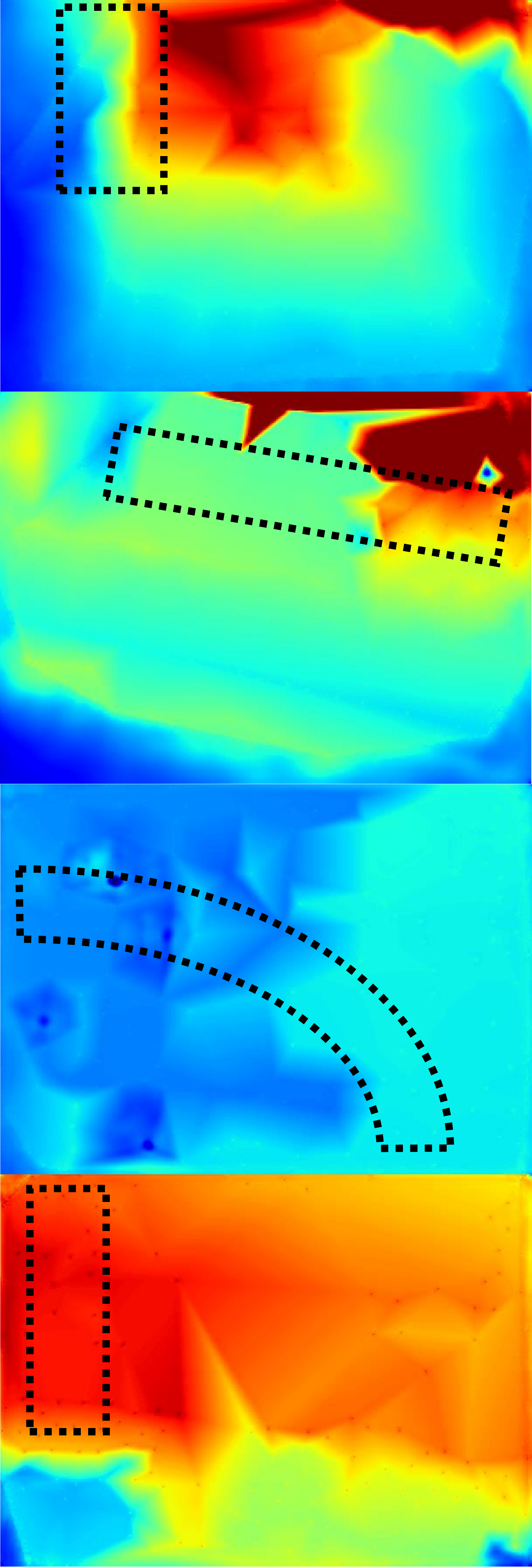}\label{fig:void_voiced}}
    \subfigure[]{\includegraphics[width=0.155\linewidth]{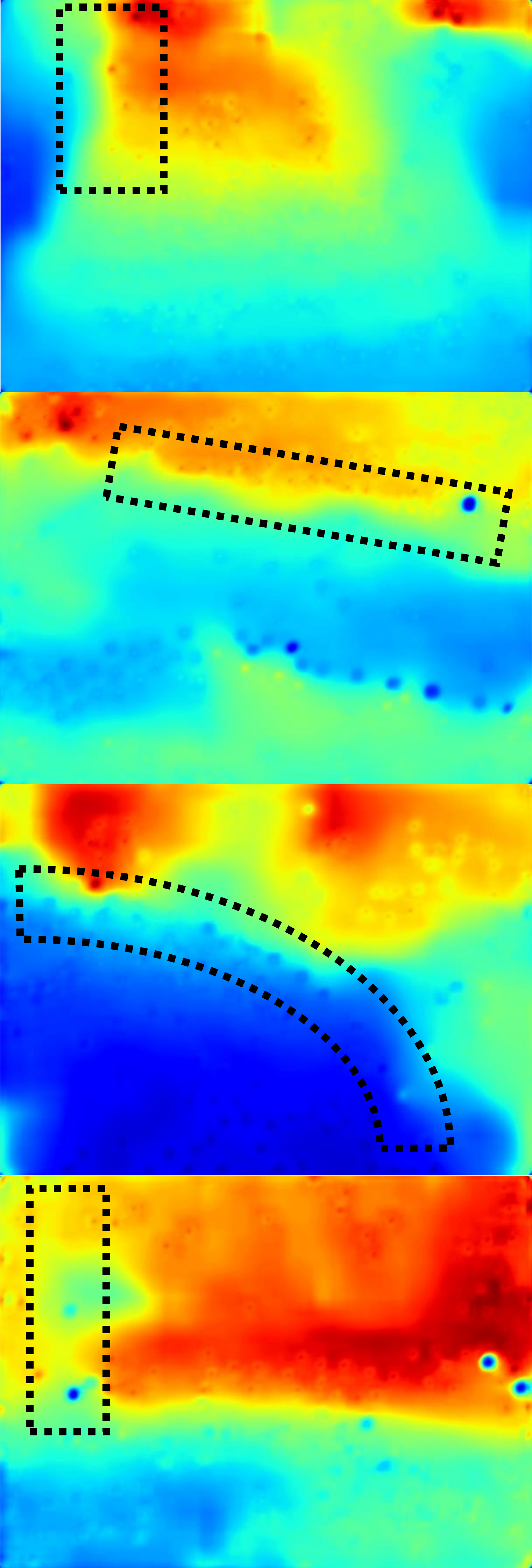}\label{fig:void_baseline}}
    \subfigure[]{\includegraphics[width=0.155\linewidth]{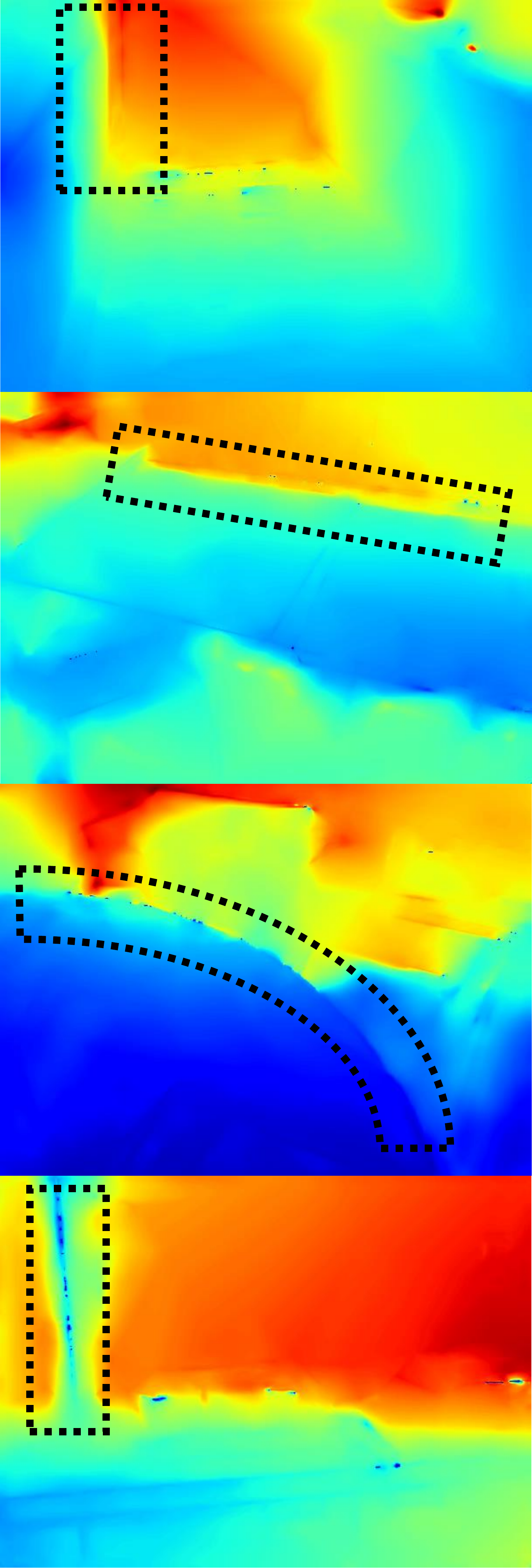}\label{fig:void_ours}}
    \subfigure[]{\includegraphics[width=0.155\linewidth]{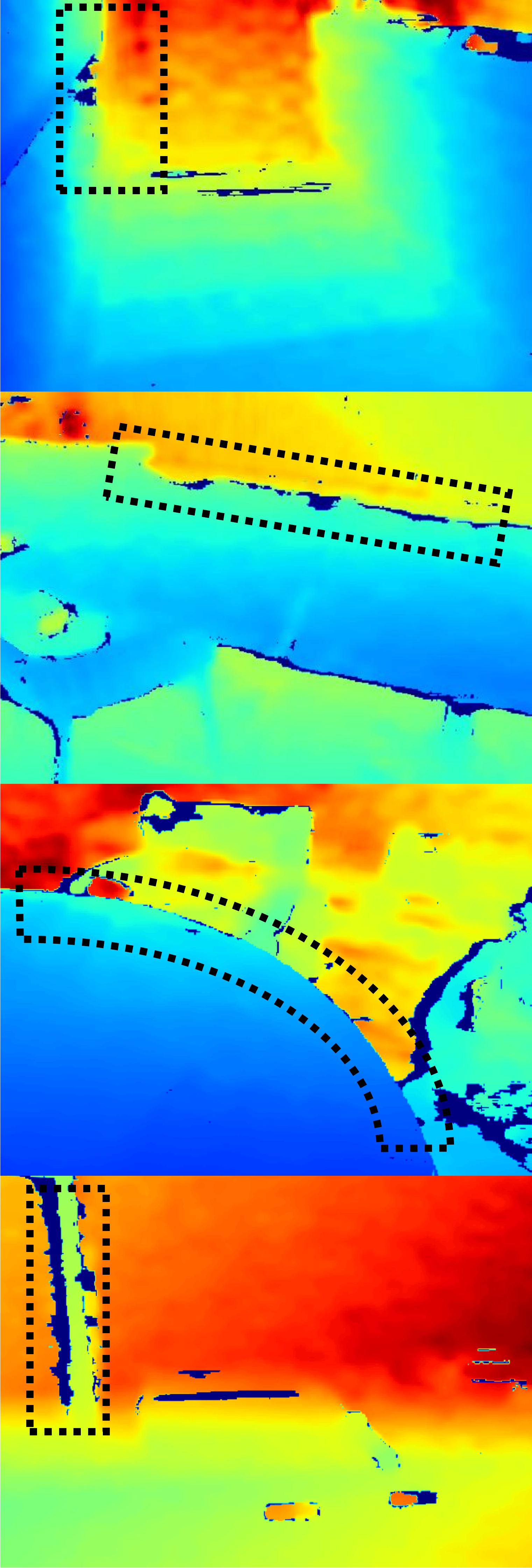}\label{fig:void_gt}}
    \caption{
        Qualitative comparison results for the VOID: 
        \subref{fig:void_point} The extracted point features, 
        \subref{fig:void_line} the extracted point and line features from UV-SLAM \cite{uvslsam}.
        Completed depth from 
        \subref{fig:void_voiced} \cite{void}, 
        \subref{fig:void_baseline} \cite{baseline}, and
        \subref{fig:void_ours} Struct-MDC.
        \subref{fig:void_gt} Ground truth depth. 
        The dashed rectangles denote the area where the improvement in the Struct-MDC stands out. 
    }
    \label{fig:void_result}
    \vspace{-0.3cm}
\end{figure}
\begin{table}[t]
\centering
\renewcommand{\arraystretch}{1.10} 
\renewcommand{\tabcolsep}{0.8mm}  
\caption{Quantitative comparison with state-of-the-arts \textcolor{color1}{methods} for the VOID \textcolor{color1}{dataset}. 
(\textit{pre}: results using their pretrained weights. 
\textit{re}: results using retrained weights.)
}

\begin{tabular}{lccccccc}
\hline
\multicolumn{2}{c}{\multirow{2}{*}{Method}} & \multicolumn{2}{c}{Error {[}mm{]} $\downarrow$} &  & \multicolumn{3}{c}{Accuracy {[}\%{]} $\uparrow$}    \\ \cline{3-4} \cline{6-8} 
\multicolumn{2}{c}{}                        & MAE                    & RMSE                   &  & $\delta_1$      & $\delta_2$      & $\delta_3$      \\ \hline
\multirow{2}{*}{\textcolor{color2}{VOICED~\cite{void}}}             & (\textit{pre}) & 167.608                & 316.017                &  & 57.499          & 74.014          & 96.974          \\
                                    & (\textit{re})  & 163.530                & 283.763                &  & 32.327          & \textbf{74.762} & 98.559          \\
\multirow{2}{*}{\textcolor{color2}{FusionNet~\cite{learning_top}}} & (\textit{pre}) & 167.600                & 346.533                &  & 55.455          & 72.721          & 97.866          \\
                                    & (\textit{re})  & 166.143                & 337.522                &  & 58.841          & 73.894          & 96.346          \\
\textcolor{color2}{KBNet (baseline)~\cite{baseline}}                    & (\textit{re})  & 143.801                & 262.643                &  & 52.286          & 69.494          & 98.724          \\
Struct-MDC (ours)                                &       & \textbf{111.332}       & \textbf{216.497}       &  & \textbf{62.074} & 74.544          & \textbf{99.003} \\ \hline
\end{tabular}
\label{table:exp_void}
\end{table}

Error metrics commonly \textcolor{color1}{adopted} for \textcolor{color1}{the} performance evaluation of depth estimation tasks \cite{baseline, error_metric3} are as follows: mean absolute error (MAE), root mean squared error (RMSE), and accuracy ratio under a threshold $\delta$ which is an absolute ratio between the estimation and the ground truth depth. 
In our experiments, overall statistic\textcolor{color1}{al analysis is based on} the MAE and RMSE, and strict thresholds $\delta_1 = 1.05$, $\delta_2 = 1.10$\textcolor{color1}{, and} $\delta_3 = 1.25^3$ are used to measure the extent of close\textcolor{color1}{ness} to the ground truth.

Struct-MDC \textcolor{color1}{used a new type of measurement: a line feature}. Therefore, depending on the dataset, the effect of line measurements is verified in various \textcolor{color1}{scenarios}.
In \textcolor{color1}{the} VOID and the NYUv2 \textcolor{color1}{datasets}, the total number of point and line features is limited to 150. 
In the PLAD \textcolor{color1}{dataset}, the point and line features estimated from the UV-SLAM are used without constraint\textcolor{color1}{s} on the total number. 

\begin{figure}[!t]
    \centering
    \subfigure[]{\label{fig:nyu_point}\includegraphics[width=0.155\linewidth]{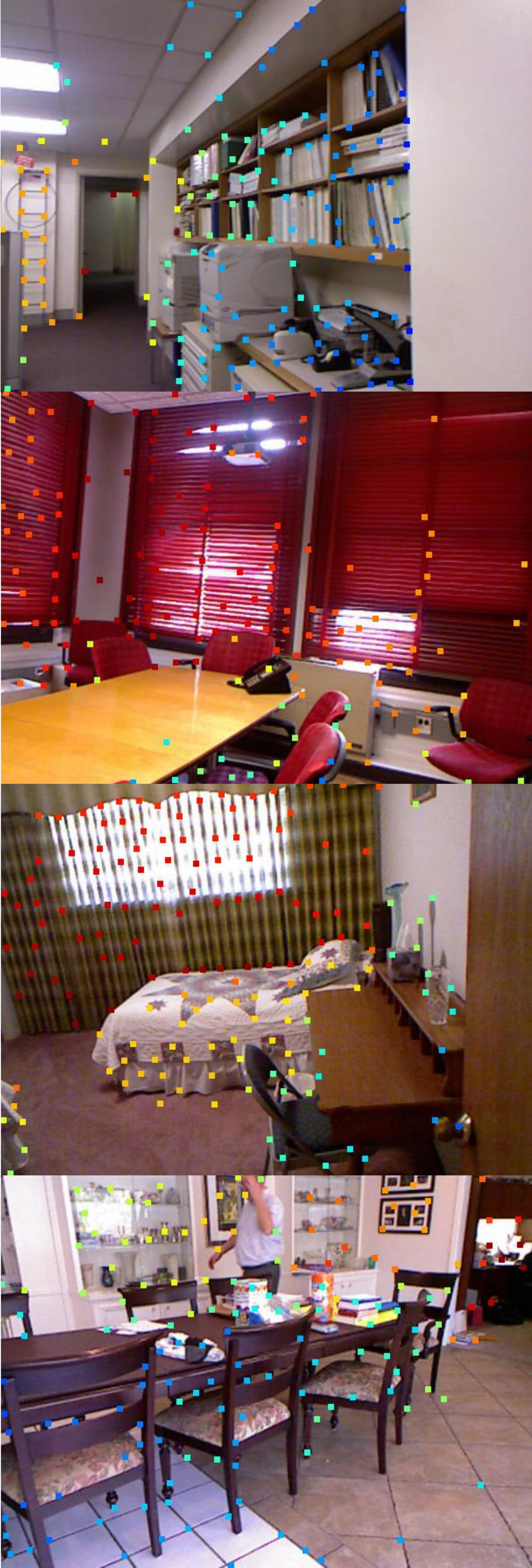}}
    \subfigure[]{\label{fig:nyu_line}\includegraphics[width=0.155\linewidth]{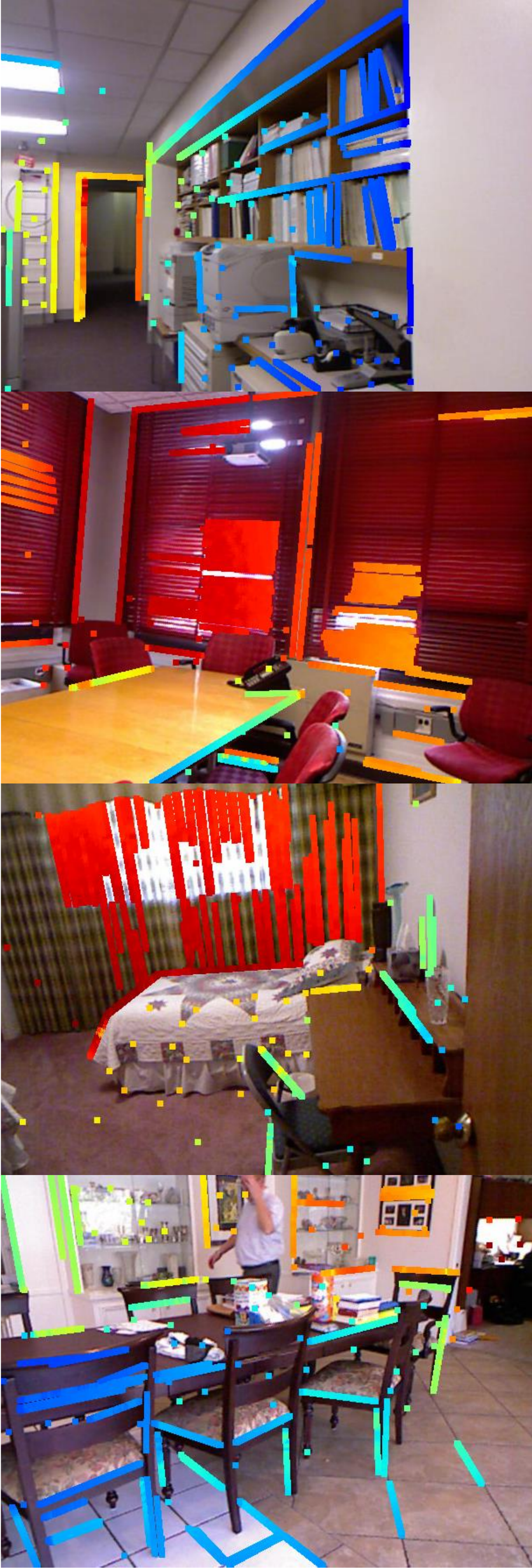}}
    \subfigure[]{\label{fig:nyu_voiced}\includegraphics[width=0.155\linewidth]{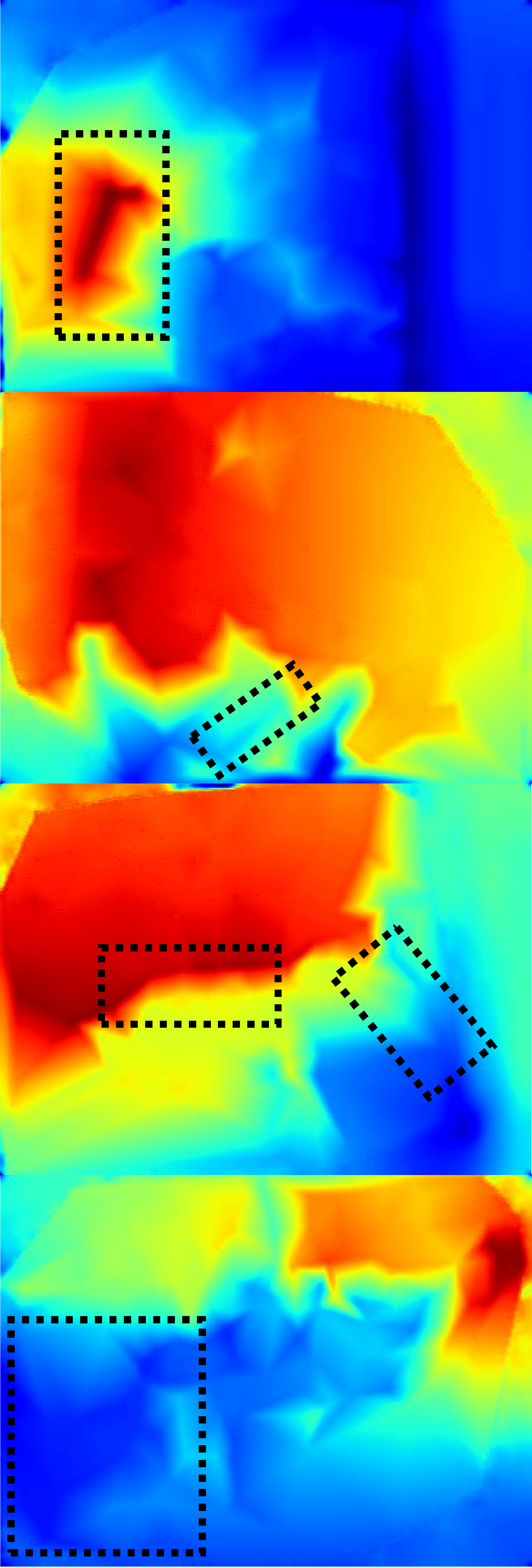}}
    \subfigure[]{\label{fig:nyu_baseline}\includegraphics[width=0.155\linewidth]{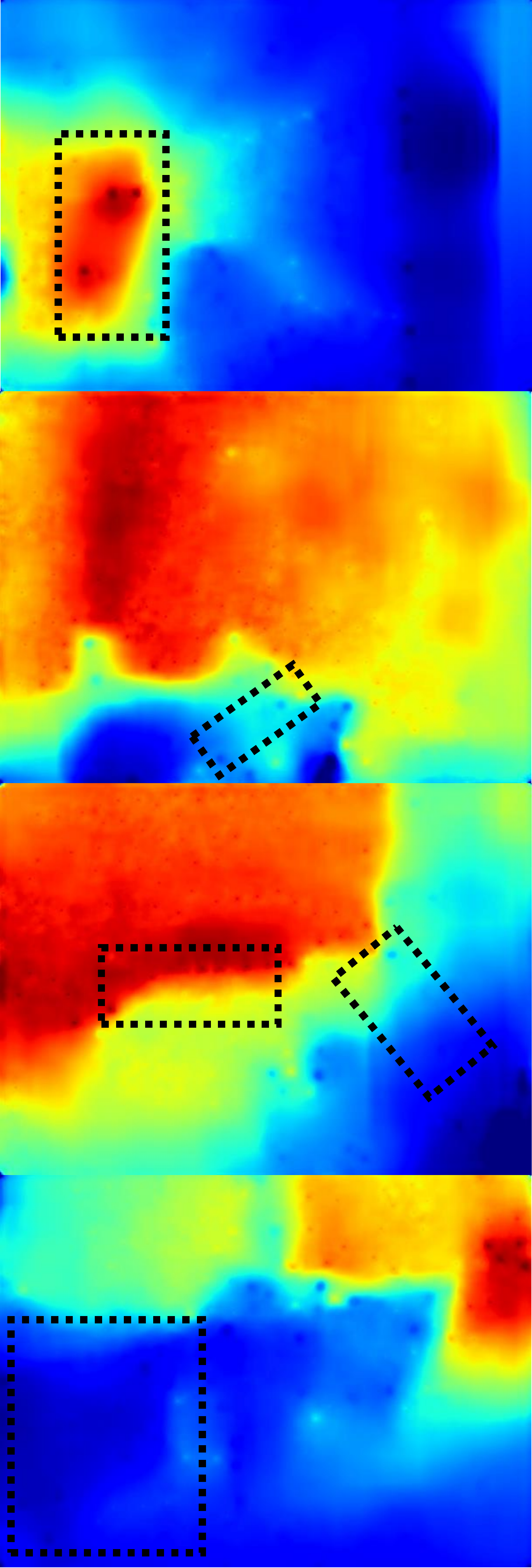}}
    \subfigure[]{\label{fig:nyu_ours}\includegraphics[width=0.155\linewidth]{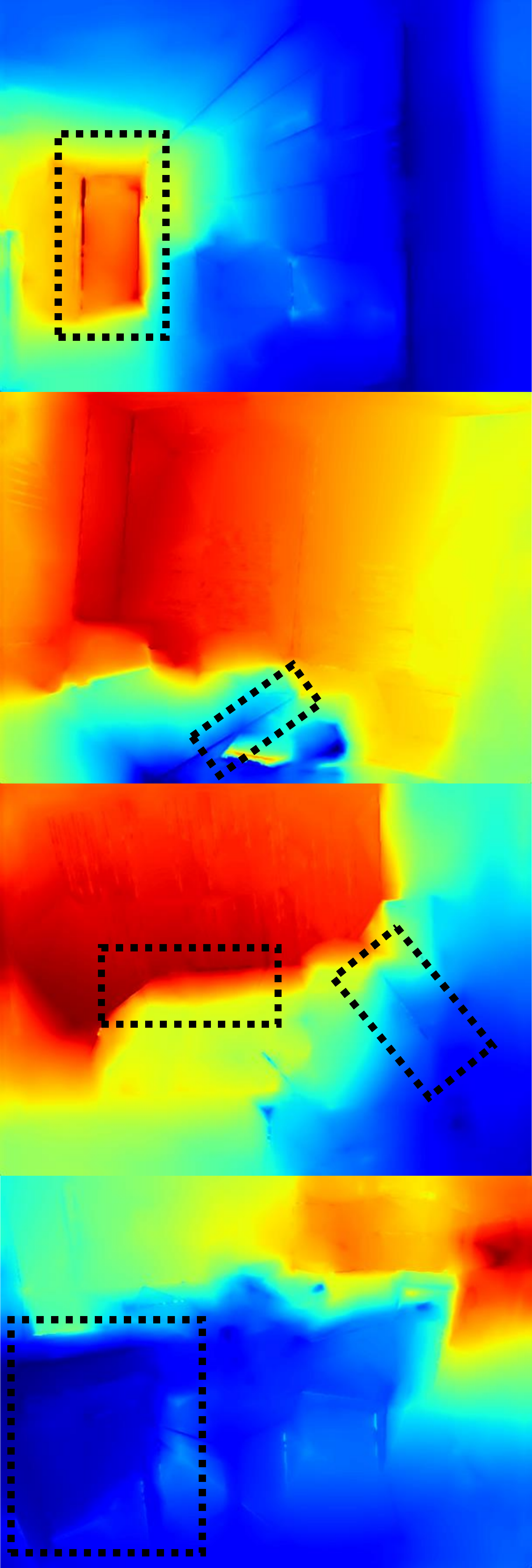}}
    \subfigure[]{\label{fig:nyu_gt}\includegraphics[width=0.155\linewidth]{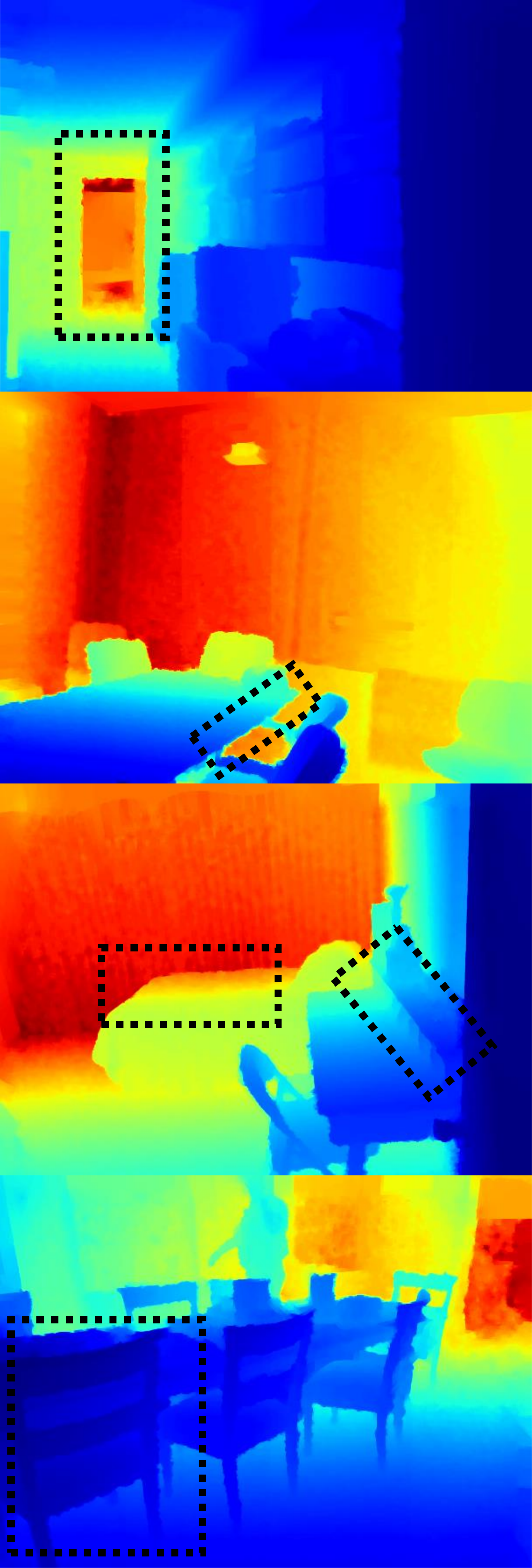}}
    \caption{
        Qualitative comparison results for the NYUv2: \subref{fig:nyu_point} The extracted point features,
        \subref{fig:nyu_line} the extracted point and line features from UV-SLAM \cite{uvslsam}.
        Completed depth from 
        \subref{fig:nyu_voiced} \cite{void}, 
        \subref{fig:nyu_baseline} \cite{baseline}, and
        \subref{fig:nyu_ours} Struct-MDC.
        \subref{fig:nyu_gt} Ground truth depth. 
        The dashed rectangles denote the area where the improvement in the Struct-MDC stands out.
    }\label{fig:nyu_result}
    \vspace{-0.3cm}
\end{figure}
\begin{table}[!t]
\centering
\renewcommand{\arraystretch}{1.10} 
\renewcommand{\tabcolsep}{0.8mm}  
\caption{Quantitative comparison with state-of-the-arts \textcolor{color1}{methods} for the NYUv2 \textcolor{color1}{dataset}.
(\textit{pre}: results using their pretrained weights. 
\textit{re}: results using retrained weights.
\textbf{U}: unsupervised, \textbf{S}: supervised
)}

\begin{tabular}{lcccccccc}
\hline
\multicolumn{3}{c}{Method}                                       & \multicolumn{2}{c}{Error {[}mm{]} $\downarrow$} & \multicolumn{1}{l}{} & \multicolumn{3}{c}{Accuracy {[}\%{]} $\uparrow$}    \\ \cline{4-5} \cline{7-9} 
\multicolumn{3}{c}{}                                             & MAE                    & RMSE                   & \multicolumn{1}{l}{} & $\delta_1$      & $\delta_2$      & $\delta_3$      \\ \hline
\multirow{2}{*}{\textcolor{color2}{VOICED~\cite{void}}}             & \multirow{2}{*}{\textbf{U}} & (\textit{pre}) & 244.416                & 407.979                &                      & 54.488          & 71.107          & 97.255          \\
                                    &                    & (\textit{re})  & 205.898                & 328.202                &                      & 56.965          & 74.747          & 99.401          \\
\multirow{2}{*}{\textcolor{color2}{FusionNet~\cite{learning_top}}} & \multirow{2}{*}{\textbf{U}} & (\textit{pre}) & 223.111                & 353.720                &                      & 51.878          & 69.801          & 99.282          \\
                                    &                    & (\textit{re})  & 208.399                & 360.653                &                      & 57.162          & 74.335          & 98.879          \\
\textcolor{color2}{KBNet (baseline)~\cite{baseline}}                    & \textbf{U}                  & (\textit{re})  & 179.817                & 297.872                &                      & 59.605          & 78.027          & 99.346          \\
Struct-MDC (ours)                                & \textbf{U}                  &       & \textbf{141.871}       & \textbf{245.548}       &                      & \textbf{67.991} & \textbf{81.999} & \textbf{99.698} \\ \hline
\textcolor{color2}{CSPN~\cite{cspn}}                                & \textbf{S}                  & (\textit{re})  & 163.152                & 245.864                &                      & 59.092          & \textbf{84.656} & \textbf{99.861} \\ \hline
\end{tabular}

\label{table:exp_nyu}
\end{table}

\subsubsection{Comparison for the VOID dataset}
In \textcolor{color1}{a} human-made environment, many line features are observed at the object boundary\textcolor{color1}{,} as shown in Fig.~\ref{fig:void_result}\subref{fig:void_line}.
The method proposed in \cite{void} depends only on \textcolor{color1}{the} depth from triangulation of point \textcolor{color1}{feature}. Therefore, as shown in Fig.~\ref{fig:void_result}\subref{fig:void_voiced}, the method \textcolor{color1}{frequently} generated incorrect triangles, estimating an ambiguous depth with mesh facet imprints. 
Our baseline, KBNet \cite{baseline}, has stains in several \textcolor{color1}{scenarios}, as shown in Fig.~\ref{fig:void_result}\subref{fig:void_baseline}. It can be \textcolor{color1}{seen} that traces of copy and paste remain in the final estimation \textcolor{color1}{because} they rely on the pooling layer to fill the low density of point features. 
In contrast, Struct-MDC does not contain any stains or incorrect imprints.
Ours employs the interpolation and MDR sequentially to fill the empty space. 
Therefore, the role\textcolor{color1}{s} of \textcolor{color1}{the two} module\textcolor{color1}{s are} adequately \textcolor{color1}{separated}, eliminating the stains and improving \textcolor{color1}{the} overall performance. 
An object boundary, which was disregarded in other methods, exists as \textcolor{color1}{a} line \textcolor{color1}{feature} and an edge of a triangle in the proposed algorithm. 
Consequently, \textcolor{color1}{such} boundaries \textcolor{color1}{are} more apparent in the final result, making the distinction between objects \textcolor{color1}{noticeable}. Remarkably, a curved boundary such as a circular table can also be estimated \textcolor{color1}{based on} our \textcolor{color1}{piecewise} linear assumption.  
The performance improvement \textcolor{color1}{can be} quantitatively \textcolor{color1}{seen from} TABLE~\ref{table:exp_void}. In Struct-MDC, the MAE and the RMSE are improved by $22.58\%$ and $17.57\%$, compared \textcolor{color1}{with} the state-of-the-art \cite{baseline}, respectively.


\subsubsection{Comparison for the NYUv2 dataset}
The qualitative comparison results are shown in Fig.~\ref{fig:nyu_result}. As in the experiments for the VOID \textcolor{color1}{dataset}, traces of triangulation remain in \cite{void} and stains are observed in \cite{baseline}. However, our methodology estimates \textcolor{color1}{a} depth close to the ground truth with the support of line features. 
In particular, Struct-MDC \textcolor{color1}{excels in} regions where point features are not extracted\textcolor{color1}{,} as shown in the first row or \textcolor{color1}{for} structural objects \textcolor{color1}{as shown in} the fourth row of Fig.~\ref{fig:nyu_result}. 
\textcolor{color1}{T}he performance improvement \textcolor{color1}{can be seen from the results in} TABLE~\ref{table:exp_nyu}. 
\textcolor{color1}{The} MAE and RMSE \textcolor{color1}{of Struct-MDC are} improved by $21.10\%$ and $17.57\%$, respectively, \textcolor{color1}{compared to those of the state-of-the-art method} \cite{baseline}. 
\textcolor{color3}{The average processing time of Struct-MDC was increased to 32.19 ms (31.07 Hz), compared to that of the baseline, 15.46 ms (64.68 Hz). However, it is still fast enough to be exploited to real robot applications.}
It is noteworthy that Struct-MDC outperforms the supervised method\textcolor{color1}{,} CSPN \cite{cspn}\textcolor{color1}{,} in \textcolor{color1}{terms of} all error metrics and \textcolor{color1}{the} accuracy metric of $\delta_1$, even though ours is an unsupervised method.

\begin{figure}[t]
    \centering
    \subfigure[]{\label{fig:plad_pl}\includegraphics[width=0.24\linewidth]{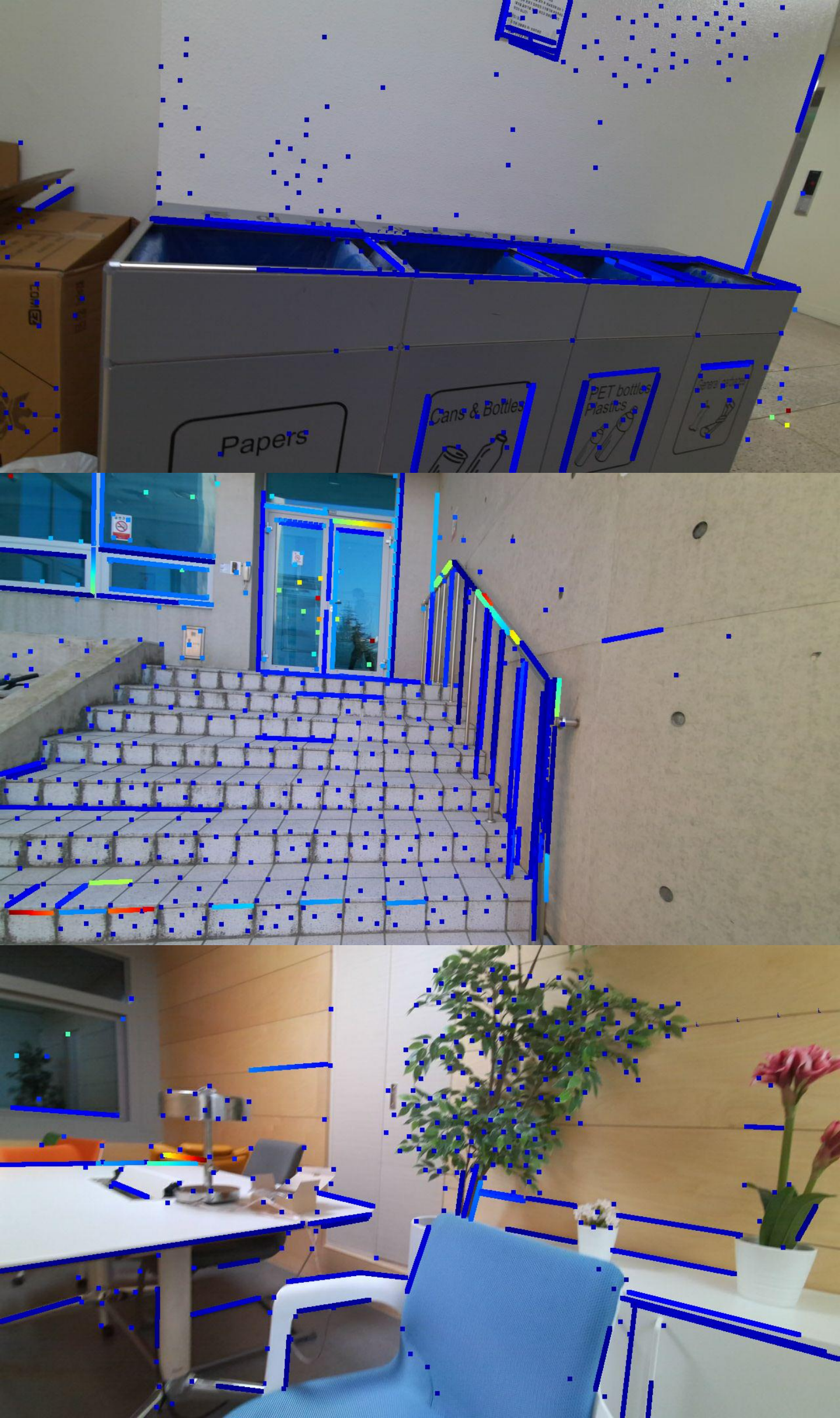}}
    \subfigure[]{\label{fig:plad_mesh}\includegraphics[width=0.24\linewidth]{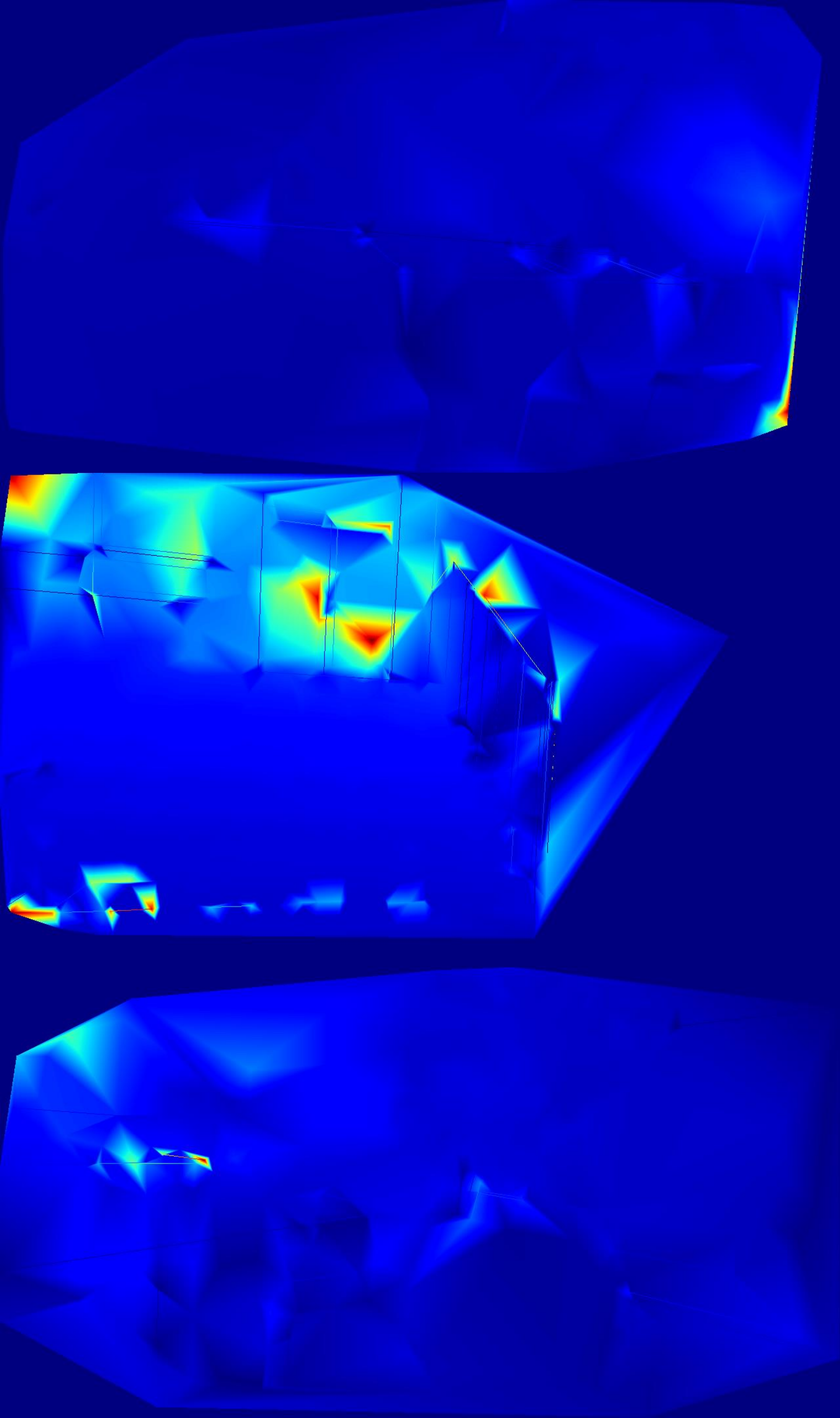}}
    \subfigure[]{\label{fig:plad_ours}\includegraphics[width=0.24\linewidth]{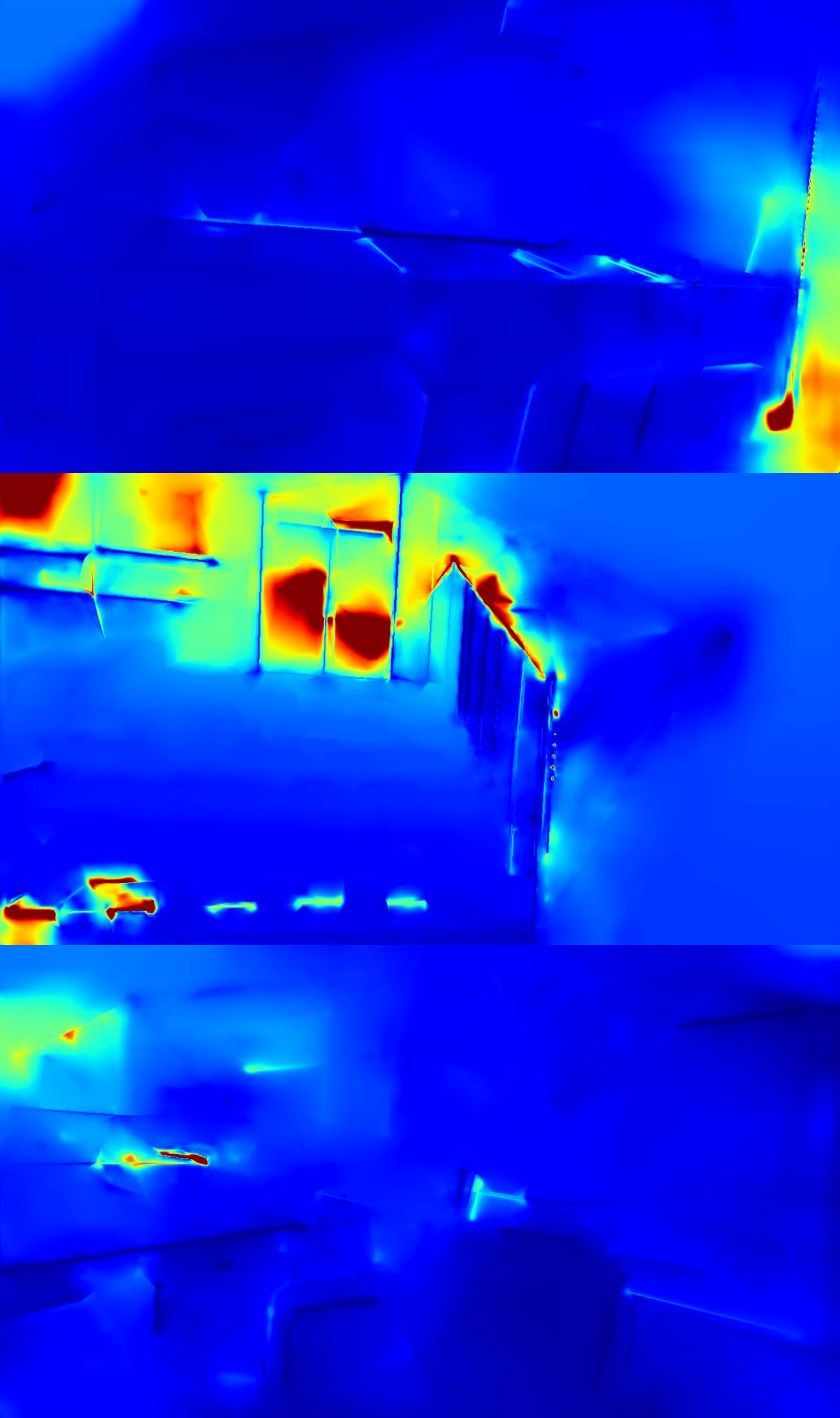}}
    \subfigure[]{\label{fig:plad_gt}\includegraphics[width=0.24\linewidth]{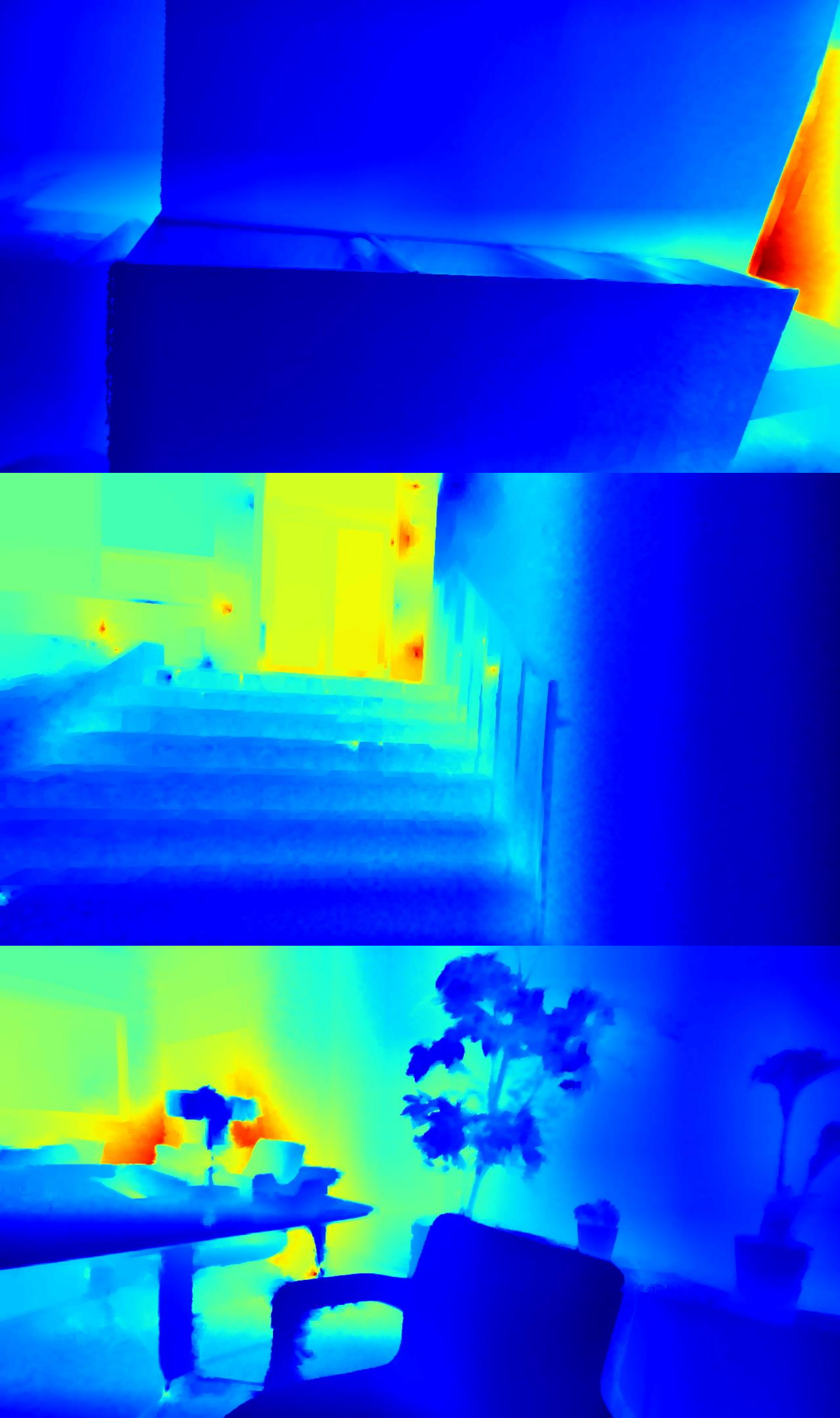}}
    \caption{Qualitative comparison results for the PLAD: 
    \subref{fig:plad_pl} point and line features from UV-SLAM \cite{uvslsam}, 
    \subref{fig:plad_mesh} interpolated mesh, 
    \subref{fig:plad_ours} completed depth from Struct-MDC, and 
    \subref{fig:plad_gt} ground truth depth.}\label{fig:plad_result}
    \vspace{-0.3cm}
\end{figure}
\begin{table}[!t]
\centering
\renewcommand{\arraystretch}{1.10} 
\renewcommand{\tabcolsep}{1.0mm}  
\caption{Quantitative comparison with state-of-the-art\textcolor{color1}{s methods} for the PLAD \textcolor{color1}{dataset}.
}
\begin{tabular}{lcccccc}
\hline
\multicolumn{1}{c}{\multirow{2}{*}{Method}} & \multicolumn{2}{c}{Error {[}mm{]} $\downarrow$} &  & \multicolumn{3}{c}{Accuracy {[}\%{]} $\uparrow$}  \\ \cline{2-3} \cline{5-7} 
\multicolumn{1}{c}{}                        & MAE                    & RMSE                   &  & $\delta_1$     & $\delta_2$     & $\delta_3$      \\ \hline
\textcolor{color2}{KBNet (baseline)~\cite{baseline}}                            & 5512.334               & 5595.194               &  & $\simeq$ 0.000 & $\simeq$ 0.000 & 9.183           \\
Struct-MDC (ours)                                        & \textbf{1170.303}      & \textbf{1481.583}      &  & \textbf{4.567} & \textbf{8.899} & \textbf{67.071} \\ \hline
\end{tabular}
%
%
\label{table:exp_plad}
\end{table}

\begin{figure*}[th!]
    \centering
    \includegraphics[width=18cm, height=5.2cm]{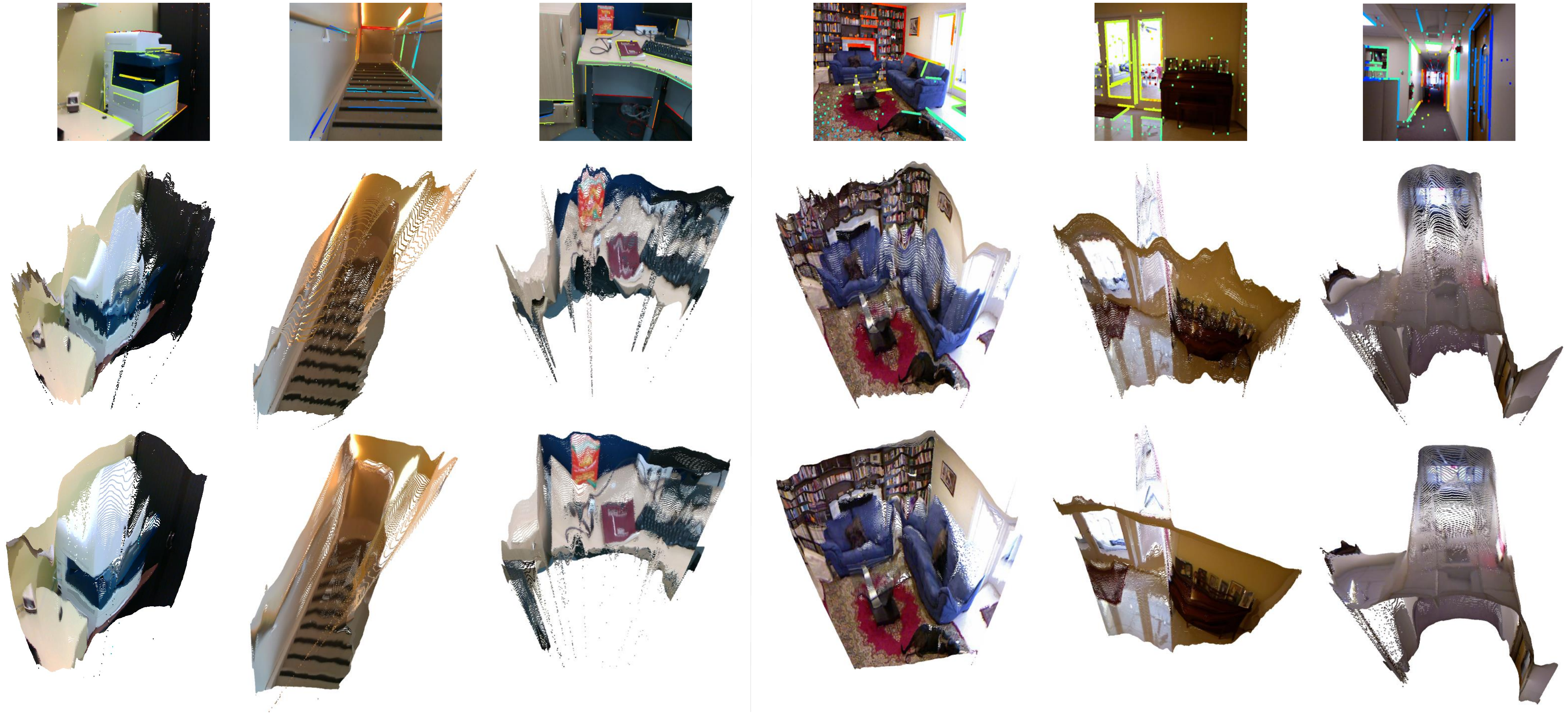}
    \caption{3D visualization of estimated depth for the VOID (left three columns) and NYUv2 (right three columns) \textcolor{color1}{datasets}: Detected features (top row), estimated depth from our baseline (middle row), and proposed method (bottom row). The proposed method \textcolor{color1}{noticeably} aligns object boundary with reduced jittering.} 
    \label{fig:3d_visualize}
    \vspace{-0.4cm}
\end{figure*} 


\subsubsection{Comparison for the PLAD dataset}
In this dataset, the spatial distribution of the features is non-uniform, and there is \textcolor{color1}{considerable} uncertainty in the estimated depth from visual SLAM. 
As \textcolor{color1}{can be inferred from} TABLE~\ref{table:exp_plad}, the baseline \textcolor{color1}{presents} deficient performance and fail\textcolor{color1}{s} to converge. \textcolor{color1}{However}, Struct-MDC \textcolor{color1}{can} successfully estimate the entire depth \textcolor{color1}{because} the empty area \textcolor{color1}{is} significantly reduced by \textcolor{color1}{the generation of}  a convex hull using point and line features, as shown in  Fig.~\ref{fig:plad_result}\subref{fig:plad_mesh}. 

The main reason \textcolor{color1}{for the failure of} the baseline is that there is no \textcolor{color1}{method} to effectively \textcolor{color1}{deal with} the no-feature area outside the convex hull. The baseline requires the assumption that at least one sparse feature \textcolor{color1}{is} included in the kernel size of the pooling layer for \textcolor{color1}{densification}. This assumption is reasonably applied inside the convex hull where sparse features exist, \textcolor{color1}{whereas} it is \textcolor{color1}{in}effective if the proportion of the no-feature areas becomes large\textcolor{color1}{r than} the overall size of the image. Actually, the estimated point features \textcolor{color1}{in the} PLAD \textcolor{color1}{dataset} are extremely sparse, $0.02\%$, compared to \textcolor{color1}{those in} the previous two datasets, $0.05\%$. 
Struct-MDC also \textcolor{color1}{partially loses its} accuracy. Nevertheless, Struct-MDC is robust to the distribution and density condition.
This is because Struct-MDC additionally utilizes the line features which compensate the sparsity of point features from visual SLAM.  
In addition, the depth can be estimated even outside the convex hull without depth information by using the MDR module.

\subsubsection{3D visualization results}
For a straightforward comparison, Fig.~\ref{fig:3d_visualize} \textcolor{color1}{shows} the 3D depth images estimated by the baseline and Struct-MDC \textcolor{color1}{on the} VOID and NYUv2 \textcolor{color1}{datasets}. 
The \textcolor{color1}{results of} the baseline \textcolor{color1}{show an} inaccurate jittering at the object boundary, \textcolor{color1}{which} is significantly reduced \textcolor{color1}{by} Struct-MDC. 
Struct-MDC can also more accurately estimate the depth inside the object than the baseline, by accurately \textcolor{color1}{capturing} the object boundary.

\subsection{Ablation Study: Mesh Depth Refinement module}\label{sec:exp_ablation}

\begin{figure}[t!]
    \centering
    \includegraphics[width=8.7cm, height=1.9cm]{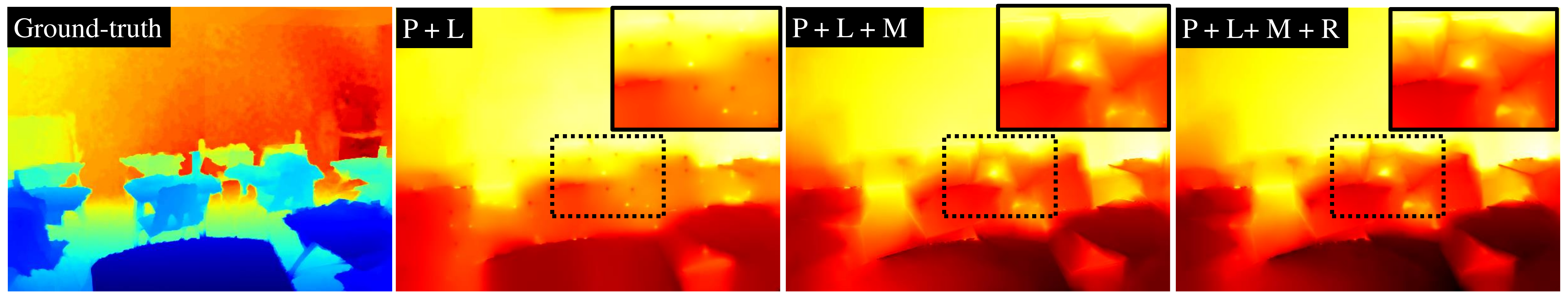}
    \caption{Error plot\textcolor{color1}{s} of different input method\textcolor{color1}{s} with ground truth depth. In three error figures on right, \textcolor{color1}{bright color implies large} error.} 
    \label{fig:error_plot}
    \vspace{-0.3cm}
\end{figure}

An ablation study was performed on two datasets to \textcolor{color1}{rigorously demonstrate the effectiveness of} the proposed MDR \textcolor{color1}{module}. 
If line features are employed, the model performance is expected to be improved as the given prior increases. Nonetheless, simply using \textcolor{color1}{a} line feature as a model input \textcolor{color1}{cannot} fully \textcolor{color1}{exploit} the \textcolor{color1}{potential of the line}. 
Fig.~\ref{fig:error_plot} and TABLE~\ref{table:ablation_1} show error plots and quantitative results according to input, respectively. P, L, M, and R denote point input, line input, mesh input, and refined mesh input, respectively.
The overall error \textcolor{color1}{is decreased} when the mesh \textcolor{color1}{is} employed \textcolor{color1}{compared to that} when simply using point and line features. The error was \textcolor{color1}{significantly} reduced when \textcolor{color1}{the} mesh refinement was performed.
When line features are simply used as an additional input to the network, blur holes still exist at the object boundary. 
The holes imply that the local smoothness loss\textcolor{color1}{,} $l_l$\textcolor{color1}{,} intensely reflect\textcolor{color1}{s} the penalty for accommodating the depth discontinuity \textcolor{color1}{in} the object boundary, and \textcolor{color1}{that} the network fell into a local minimum with a simple copy and paste. \textcolor{color1}{Because} the baseline employs a pooling layer for \textcolor{color1}{densification, and} the model \textcolor{color1}{is prone to} fall into this local minimum. 
In contrast, blur holes \textcolor{color1}{are} removed with the interpolated mesh depth as \textcolor{color1}{the} input\textcolor{color1}{, whereas} the object boundary \textcolor{color1}{is} captured sufficiently. \textcolor{color1}{This is because} the empty depth is filled \textcolor{color1}{by} interpolation with awareness of the object boundary \textcolor{color1}{by} CDT. 
The advantage of using \textcolor{color1}{a} mesh \textcolor{color1}{emerges} if the proposed MDR module is utilized jointly. Compared \textcolor{color1}{to} the case where point and line features \textcolor{color1}{are simply} used in the baseline model (P + L), the proposed methodology (P + L + M + R) dramatically improve\textcolor{color1}{s} the overall performance. 

%
\begin{table}[t!]
\centering
\renewcommand{\arraystretch}{1.1} 
\renewcommand{\tabcolsep}{1.1mm}  
\caption{Ablation study: Effectiveness of \textcolor{color1}{p}roposed \textcolor{color1}{m}esh \textcolor{color1}{d}epth \textcolor{color1}{r}efinement module. 
(P: point feature, L: line feature, M: interpolated mesh, R: \textcolor{color1}{p}roposed refinement module.
)}
\begin{tabular}[ht]{clcccccc}
\hline
\multirow{2}{*}{Dataset}                   & \multicolumn{1}{c}{\multirow{2}{*}{Method}} & \multicolumn{2}{c}{Error {[}mm{]} $\downarrow$} &  & \multicolumn{3}{c}{Accuracy {[}\%{]} $\uparrow$}    \\ \cline{3-4} \cline{6-8} 
                                          & \multicolumn{1}{c}{}                        & MAE                    & RMSE                   &  & $\delta_1$      & $\delta_2$      & $\delta_3$      \\ \hline
\multirow{4}{*}{VOID}                      & P (baseline)                                & 143.801                & 262.643                &  & 52.286          & 69.494          & 98.724          \\
                                          & P + L                                       & 124.243                & 235.622                &  & 58.642          & 72.745          & 98.949          \\
                                          & P + L + M                                   & 120.917                & 228.467                &  & 58.741          & 73.418          & 99.002          \\
                                          & P + L + M + R                               & \textbf{111.332}       & \textbf{216.497}       &  & \textbf{62.074} & \textbf{74.544} & \textbf{99.003} \\ \hline
\multicolumn{1}{l}{\multirow{4}{*}{NYUv2}} & P (baseline)                                & 179.817                & 297.872                &  & 59.605          & 78.027          & 99.346          \\
\multicolumn{1}{l}{}                       & P + L                                       & 163.618                & 270.276                &  & 62.230          & 79.185          & 99.500          \\
\multicolumn{1}{l}{}                       & P + L + M                                   & 144.853                & 248.293                &  & 67.381          & 81.728          & 99.687          \\
\multicolumn{1}{l}{}                       & P + L + M + R                               & \textbf{141.871}       & \textbf{245.548}       &  & \textbf{67.991} & \textbf{81.999} & \textbf{99.698} \\ \hline
\end{tabular}
\label{table:ablation_1}
\vspace{-0.4cm}
\end{table}

%% file: 5.conclusion.tex
\vspace{-0.2cm}

\section{Conclusions} \label{sec:cons}

In summary, this study introduced line features from visual SLAM as new measurements to \textcolor{color1}{a} depth completion task for the first time.
For this, a three-step approach to efficiently use line features: structural sketch, refinement, and estimation\textcolor{color1}{,} was proposed with a bridge module between the conventional and deep learning approaches. 
Struct-MDC \textcolor{color1}{was} proven to achieve state-of-the-art performance \textcolor{color1}{based on} experiments with various datasets. 
Furthermore, \textcolor{color1}{an} ablation study proved that the CDT mesh with the MDR module \textcolor{color1}{significantly increases} the overall performance. 
Line features have strong potential in that it captures the depth discontinuity at \textcolor{color1}{the} object boundary, \textcolor{color1}{which is} a \textcolor{color1}{continuing} deep learning problem. 
Nevertheless, depth completion utilizing line feature\textcolor{color1}{s} from visual SLAM is an area that has not been actively explored. Therefore, to contribute to this society, we disclose our code and \textcolor{color1}{the} custom dataset\textcolor{color1}{,} PLAD. 
\textcolor{color3}{The limitation of Struct-MDC is that the average processing time was slightly increased compared to the baseline.} 
We will optimize the model for future works to reduce the processing time and apply it to real robot exploration.



\vspace{-0.4cm}